\documentclass[journal]{IEEEtran}
\newtheorem{definition}{Definition}
\newtheorem{proposition}{Proposition}

\ifCLASSINFOpdf
\else
\fi
\usepackage[flushleft]{threeparttable} %
\usepackage{booktabs,caption,verbatim}
\usepackage{stfloats}
\usepackage{mathrsfs}
\usepackage{graphicx}
\usepackage[utf8]{inputenc}
\usepackage[english]{babel}
\usepackage{tikz,xcolor,hyperref}
\usepackage{marvosym}
\usepackage{amsmath}
\usepackage{amssymb}
\usepackage{enumerate}
\usepackage{lscape}
\usepackage{adjustbox}
\usepackage{xcolor,colortbl}
\usepackage{multirow,tabularx}
\usepackage{bm}

\definecolor{Gray}{gray}{0.85}
\newcolumntype{a}{>{\columncolor{Gray}}c}

\definecolor{lime}{HTML}{A6CE39}
\DeclareRobustCommand{\orcidicon}{%
	\begin{tikzpicture}
	\draw[lime, fill=lime] (0,0) 
	circle [radius=0.16] 
	node[white] {{\fontfamily{qag}\selectfont \tiny ID}};
	\draw[white, fill=white] (-0.0625,0.095) 
	circle [radius=0.007];
	\end{tikzpicture}
	\hspace{-2mm}
}

\foreach \x in {A, ..., Z}{%
	\expandafter\xdef\csname orcid\x\endcsname{\noexpand\href{https://orcid.org/\csname orcidauthor\x\endcsname}{\noexpand\orcidicon}}
}

\newcommand{\treviewerone}[1]{{\color{black}{#1}}}
\newcommand{\treviewertwo}[1]{{\color{black}{#1}}}
\newcommand{\treviewerthree}[1]{{\color{black}{#1}}}
\newcommand{\treviewerfor}[1]{{\color{black}{#1}}}

\begin{document}
	%
	\title{Analyzing Dominance Move (MIP-DoM) Indicator for Multi- and Many-objective Optimization}
	%
	%
	%
	
	\author{
		
		Claudio Lucio do Val Lopes\orcidA{},
		\thanks{Corresponding author: Cláudio Lúcio V. Lopes is with Postgraduate Program in Mathematical and Computational Modeling, CEFET-MG, Brazil (email: claudiolucio@gmail.com)}
		Flávio Vinícius Cruzeiro Martins \orcidB{},
		\thanks{Flávio V. C. Martins is with the the Computer Department at the Centro Federal de Educação Tecnológica de Minas Gerais, 30510-000 Brazil (e-mail: flaviocruzeiro@cefetmg.br).}
		Elizabeth Fialho Wanner \orcidC{},
		\thanks{Elizabeth F. Wanner is with the Computer Department at the Centro Federal  de  Educação Tecnológica de Minas Gerais, 30510-000 Brazil  (e-mail: efwanner@cefetmg.br).}
		and~Kalyanmoy Deb~\IEEEmembership{Fellow,~IEEE} \orcidD
		\\[3mm]
		\thanks{Kalyanmoy Deb is with the Department of Computer Science and
			Engineering, Michigan State University, East Lansing, MI 48824 USA (email: kdeb@msu.edu).}
	}

	\maketitle
	
	\begin{abstract}
		Dominance move (DoM) is a binary quality indicator that can be used in multi-objective and many-objective optimization to compare two solution sets obtained from different algorithms. The DoM indicator can differentiate the sets for certain important features, such as \textit{convergence}, \textit{spread}, \textit{uniformity}, and \textit{cardinality}. DoM \treviewerthree{does not use any reference, and it} has an intuitive and physical meaning, similar to the $\epsilon$-indicator, and calculates the minimum total move of members of one set
		\treviewertwo{so that all elements in another set are to be dominated or identical to at least one member of the first set.}
		Despite the aforementioned properties, DoM is hard to calculate, particularly in higher dimensions. There is an efficient and exact method to calculate it in a bi-objective case only. This work proposes a novel approach to calculate DoM using a mixed integer programming (MIP) approach, which can handle sets with three or more objectives and is shown to overcome the $\epsilon$-indicator's information loss. Experiments, in the bi-objective space, are done to verify the model's correctness. Furthermore, other experiments, using 3, 5, 10, 15, 20, 25 and 30-objective problems are performed to show how the model behaves in higher-dimensional cases. Algorithms, such as IBEA, MOEA/D, NSGA-III, NSGA-II, and SPEA2 are used to generate the solution sets (however any other algorithms can also be used with the proposed MIP-DoM indicator). 
		Further extensions are discussed to handle certain idiosyncrasies with some solution sets and also to improve the quality indicator and its use for other situations.
	\end{abstract}
	

	\section{Introduction}
	\label{intro}
	\IEEEPARstart{P}{roblems} with conflicting objectives arise in most real-world optimization problems. These problems are solved using evolutionary multi-objective (EMO) or many-objective optimization (EMaO) techniques \cite{Chugh:2019:SHC:3333833.3333883} to find a set of Pareto-optimal solutions in three or less, or more than three-objective problems, respectively. EMO/EMaO algorithms are applied in different domains \cite{Gonzlezlvarez2013AnalysingTS,Deb_Kalyam_2008}. The solution sets for these problems are formed in such a way that each solution represents a {\em trade-off\/} among the conflicting objectives. 
	
	In contrast to the single-objective optimization, the solutions generated by EMO/EMaO algorithms can be difficult to obtain and compare. Such difficulty grows as the need for having an increasingly large number of candidate solutions with an increase in number of objectives. When there are two or three objectives only, some graphical techniques help to examine the solution set visually. However, when the number of objectives is greater than three, this task is challenging, needing more advanced visualization techniques that can present location, shape, and solution set distribution   \cite{10.1145/2792984,DBLP:journals/swevo/IbrahimRMD18,Bhattacharjee2020,paletteviz}.
	
	Quality indicators are suitable for situations when we need to compare two or more trade-off solution sets \cite{ecki-perf}. They have been used to compare the outcomes of multi-objective algorithms quantitatively. In a recent paper \cite{Li:2019:QES:3320149.3300148}, 100 quality indicators were discussed considering some state-of-the-art indicators focusing on which quality aspects these indicators have, as well as their strengths and weaknesses.
	
	In \cite{DBLP:journals/corr/LiY17a}, a new quality measure, called dominance move (DoM), was proposed. This measure is able to capture different aspects of a solution set's quality, such as spread, uniformity, and cardinality. Moreover, the indicator is also Pareto compliant \cite{ecki-perf}, meaning that a move from a set to another completely dominated set will have a smaller indicator value than the other way around. 
	The DoM measures the minimum `effort' that one solution set has to make 
	\treviewertwo{to all elements of a second set are dominated or become identical to at least one element of the first set}, 
	more specifically the absolute sum of the movements (i.e., Minkowski distance) in objective directions needed to make the \treviewertwo{above-mentioned moves happen}.

	\treviewertwo{DoM does not require any pre-defined set of points, such as a reference point or a reference set. It does not demand any normalization of the objective functions, and it is not affected by dominance resistant solutions.}\treviewerthree{ DoM takes advantage of all available information, unlike the $\epsilon$-indicator which uses information from one particular solution in the set.} Authors to the original study presented an exact approach to calculate DoM for the bi-objective case. Despite the low computational cost in the bi-objective case, the exact DoM computational approach cannot be applied directly or extended to problems with three or more objectives.  
	
	\treviewertwo{Although DoM presents good quality aspects as an indicator,  its calculation for problems having three or more objective functions is unknown  so far \cite{Li2015, Li2015Thesis, Li:2019:QES:3320149.3300148}. Its application to bi-objective problems only limits its spread into the research area. }
	
	In \cite{lopes2020assignment}, as an attempt to circumvent the bi-objective limitation, authors proposed an assignment problem formulation to calculate the DoM in problems with three or more objective functions. They carried out preliminary experiments using problems with three objectives. Results showed that the assignment formulation for DoM is valid, but the cardinality of the solution sets imposed a handicap in preventing its use in solutions sets with more than 20 solution candidates. {\color{black}{In this paper, we facilitate DoM computation with a mixed-integer linear programming (MIP) formulation for its application to a large solution set and also evaluate MIP-DoM procedure in a comprehensive manner.}}
	
	This work focuses on a DoM calculation procedure based on a mixed integer programming (MIP) approach \cite{DBLP:books/daglib/0023873} aiming to overcome this difficulty. The mixed integer formulation is presented here. Initial experiments on two-objective problems amply show that the MIP-DoM formulation is correct. More specifically, this paper presents the following contributions: (i) the proposed MIP model for DoM calculation is valid for any objective space dimension and cardinality of the supplied solution sets, (ii) extensive evaluation of the proposed MIP-DoM formulation and approach in a number of common problem sets having 3, 5, 10, 15, 20, 25 and {\color{black}30} objective functions, (iii) comparison with optimal DoM solutions in cases where it is possible to achieve the optimal solution, and (iv) raise of questions and details about the model behaviour for some test sets, including future research paths to tackle other problems.
	
	This paper is organized as follows. In Section \ref{definitions}, some problems and definitions are posed, regarding unary, binary, and $k$-ary quality indicators. Some well-known quality indicators are presented, as well. Dominance Move (DoM), its definition, properties, and features are presented in Section \ref{sec:dom}. Our mixed-integer approach to calculate DoM is presented in Section \ref{MIPDOM}. The MIP-DoM model is introduced with some comments and considerations. Experiments are discussed in Section \ref{experiments}, and it starts with bi-objective case presenting the correctness and model validity. Following, some common multi-objective problem sets and algorithms are used to compare MIP-DoM with the $\epsilon$-indicator, $IGD^+$, and Hypervolume indicators. Still in this section, the advantages of DoM over the $\epsilon$-indicator are presented, although they share a similar interpretation regarding its use. Still in Section \ref{experiments}, the many-objective experiments are discussed using problem sets with 5, 10, 15, 20, 25 and {\color{black}30} objectives. \treviewerfor{In Section \ref{sec:further}, further extensions are discussed to handle certain idiosyncrasies with some solution sets and also to improve the quality indicator and its use for other situations.} Finally, in Section \ref{conclusion}, some final considerations, recommendations, and future research paths are presented.
	
	\newcommand{\boldx}{\mbox{${\mathbf x}$}}
	\newcommand{\boldz}{\mbox{${\mathbf z}$}}
	\newcommand{\boldy}{\mbox{${\mathbf y}$}}
	
	\section{Problems and definitions}\label{definitions}
	
	In general, a multi-objective optimization problem (MOP) includes a decision variable vector, \boldx, from a feasible decision space $\Omega \subseteq \mathbb{R}^{N}$, and a set of $M$ objective functions. Without loss of generality, the minimization of an  MOP can be simply defined as \cite{Yuan2018}:
	\begin{align}\label{MOP_definition}
	\mbox{Minimize} \quad F(\boldx) = {[f_{1}(\boldx), \ldots, f_{M}(\boldx)]}^{T}, \quad  \boldx \in \Omega.
	\end{align}
	$F: \Omega \rightarrow \Theta \subseteq \mathbb{R}^{M}$ is a mapping from the feasible decision space $\Omega$ to vectors in the $M$-dimensional objective space $\Theta$. We are interested in the evaluation of these objective vector (solution) sets, and the comparison relation among them.
	
	{\color{black}\begin{definition}{\textit{Dominance}:} 
			Consider two solutions, \bm{$p$}, \bm{$q$}  $\in \Theta$. \bm{$p$} is said to \textit{dominate} \bm{$q$} if \treviewertwo{{$p_m$} $\leq$ $q_m$} for $1 \leq m \leq M$ \treviewertwo{and $p_m<q_m$ for at least one $m$.} It is denoted as  \bm{$p$} $\prec$ \bm{$q$}. 
	\end{definition}}
	A solution \bm{$p$} $\in \Theta$ is called {\color{black}{\em Pareto-optimal\/}} (or {\em efficient\/}), if there is no \bm{$q$}  $\in  \Theta$ that dominates \textit{\bm{$p$}}. 
	Such solutions constitutes a Pareto-optimal front (or an efficient set) in the objective space. 
	
	\treviewertwo{Note that a solution cannot dominate itself, that is, if $\bm{q}=\bm{p}$, $\bm{p}$ does not dominate $\bm{q}$, and vice versa. But if $p_j=q_j-\epsilon_j$ for a specific $j$ and $\epsilon_j\rightarrow 0$, and for all other objectives ($i=1,2,\ldots,M$ and $i\ne j$) $p_i = q_i$, then $\bm{p}$ $\prec$ \bm{$q$}.}  

	The comparison between two or more solution sets is of great importance in EMO algorithms and hence a considerable amount of efforts have been made to define different performance indicators. The indicators can be used to compare the outcomes of multi-objective algorithms or even assist an algorithm during the search process for non-dominated candidates. It is paramount to make more precise statements when applying a quality indicator comparison, for example, if one algorithm is better than another, how much better is it? The following definition formalizes the quality indicators \cite{ecki-perf}:
	\treviewertwo{The above definitions can be extended to compare two sets $\bm{P}$ and $\bm{Q}$. 
		\begin{definition}{\textit{Dominance of Sets}:} 
			A set $\bm{P}$ is said to dominate another set $\bm{Q}$, if every member of $\bm{Q}$ gets dominated by at least one member of $\bm{P}$. It is denoted as $\bm{P} \prec \bm{Q}$.
		\end{definition}
		Such comparisons can be used to definite quality indicators.}
	\begin{definition}{\textit{Quality indicator}:} 
		\treviewertwo{A \textit{k}-ary} quality indicator \textit{I} is a function \textit{I}:${\Theta}^k \rightarrow \mathbb{R}$, 
		which assigns each vector of \textit{k} solution sets {(\bm{$P_1$}, \bm{$P_2$}, \ldots, \bm{$P_k$})} a real value I{(\bm{$P_1$}, \bm{$P_2$} ,\ldots, \bm{$P_k$})}.
	\end{definition}
	
	The quality indicators can be unary, binary, or \textit{k}-ary, defining a value to one solution set, two solution sets, or \textit{k} solution sets, respectively.
	In \cite{Li:2019:QES:3320149.3300148}, 100 indicators were listed and some were  discussed in detail. In this work, four properties of indicator quality are analyzed: \textit{convergence}, \textit{spread}, \textit{uniformity}, and \textit{cardinality}. 
	%
	It is expected that the dominance must be a central criterion in reflecting the \textit{convergence} of solution sets \cite{ecki-perf}. For two solution sets \bm{$P$} and \bm{$Q$}, for example, if \bm{$P$} dominates \bm{$Q$}, then $I(\bm{P}, \bm{Q}) = 0$. \treviewertwo{If $\bm{P}$ dominates some points of $\bm{Q}$, and $\bm{Q}$ does not dominate any point of $\bm{P}$, it is reasonable to expect that the indicator value $I(\bm{P}, \bm{Q})$ is better than $I(\bm{Q} , \bm{P})$.}

	The \textit{spread} of a solution set must consider the region that the set is covering. It involves finding the extreme points and a well-distributed set of points in the interior of the Pareto set \cite{deb-bbok-01}. 
	
	The number of solutions in the set is another property, known as \textit{cardinality} (in general, solutions sets with more candidates and generated with the same computational resources are preferred).  Finally, a good indicator must prefer a set with uniformly distributed points, \textit{uniformity}, showing an equidistant spacing (either along the Pareto surface or in an Euclidean sense) amongst solutions.
	
	It is plausible to add more details about the indicators due to their usage, such as the computational cost. Some indicators present all quality aspects but are hard to compute, notably in high dimensions or for high-cardinality sets. Other indicator details also deserve to be mentioned, such as the necessity for reference point or set, additional parameters, how to deal with scale, and normalization, etc., but we do not consider them in this study.  
	
	There are many indicators available, and they have been used in numerous situations in the literature \cite{Li:2019:QES:3320149.3300148}. Hypervolume (HV) \cite{8625504,Yang2019,Bradford_2018}, generational distance (GD) \cite{10.1007/978-3-319-15892-1_8}, inverted generational distance plus (IGD+) \cite{10.1007/978-3-319-15892-1_8}, KKTPM \cite{7283599}, and $\epsilon$-indicator \cite{ecki-perf} are some examples. Our supplementary material further present the definition of some indicators.
	
	
	Hypervolume has been widely applied in the evolutionary community as a unary indicator choice \cite{Audet2018PerformanceII}. It is still important to note its drawbacks:  the increasing exponential cost (due to the number of objectives) and the necessity to designate an explicit reference point \cite{Ishibuchi_Ryo}. The reference point affects the ordering of pairs of incomparable sets. Some community efforts have focused on the discovery of new indicators, which can overcome hypervolume's limitations while providing a good set of properties.
	
	An ideal quality indicator must present the four facets. Additionally, it must have a low computational cost, and it does not need a normalization (due to objective scale) and any additional parameters or reference points/sets.

	\section{Dominance Move}\label{sec:dom}
	
	Dominance move (DoM) is an intuitive indicator that is commensurate with all four desirable properties \cite{DBLP:journals/corr/LiY17a}. The first idea presenting DoM came from the performance comparison indicator (PCI) \cite{Li2015}. Examining the PCI proposal, it is quite similar to DoM in its essential purpose. PCI, a binary quality indicator, builds up a reference set using two solution sets, \bm{$P$} and \bm{$Q$}. This reference set is then split up into clusters, and the indicator calculates the movement distance {\color{black}for one set to dominating the other set}. 
	
	Dominance move is a measure for comparing two sets of multi-dimensional points, being classified as a binary indicator. It considers the minimum overall movement of points in one set needed to \treviewertwo{dominate each and every member of the other set}. 
	
	
	\treviewertwo{\begin{definition}{Dominance Move (DoM): }
			Consider that \bm{$P$} and \bm{$Q$} are two sets of points, with  $\bm{p}_i$ points $i \in \{{1,\ldots,|\bm{P}|}\}$ and $\bm{q}_j$ points $j \in \{{1,\ldots,|\bm{Q}|}\} $. The dominance move of \bm{$P$} to \bm{$Q$}, $DoM(\bm{P},\bm{Q})$, is the minimum total distance of moving points of \bm{$P$}, such that 
			the moved set ${\bm{P'} =  \{{\bm{p}_1'}, {\bm{p}_2'}, \ldots, {\bm{p}_{|\bm{P}|}'} \}}$ (with some or all $\bm{p}_i'$ are allowed to be infeasible) from  ${\bm{P} = \{\bm{p}_1, \bm{p}_2, \ldots, \bm{p}_{|\bm{P}|}\}}$ dominates \bm{$Q$} and that the total move from $\bm{P}$ to $\bm{P'}$ must be minimum \cite{DBLP:journals/corr/LiY17a}.
	\end{definition}}
	\treviewertwo{It is important to highlight that above definition is similar in concept to that in \cite{DBLP:journals/corr/LiY17a}, except that an inconsistent definition of {\em weakly} dominance operator compared to that in the multi-objective optimization literature \cite{miet} was used. However, our dominance move definition is adequate with the standard Pareto dominance concept and achieves both the desired concepts originally proposed:
		\begin{enumerate}
			\item Every element of $\bm{Q}$ gets dominated by at least one element of $\bm{P}$. 
			\item Since DoM corresponds to the minimum move distance, one of the moved elements of $\bm{P'}$ can be vanishingly better ($\epsilon \rightarrow 0$) in one or more objectives compared to a dominating element of $\bm{Q}$, thereby providing a vanishingly close DoM value compared to the original DoM. 
	\end{enumerate}}
	
	The formal expression of DoM can be stated as: 
	\begin{equation}\label{DPQ}
	DoM(\bm{P},\bm{Q}) = \underset{{\color{black}{\bm{P}' \prec  \bm{Q}}}} {\min} \sum\limits_{i=1}^{|\bm{P}|} d(\bm{p}_i,\bm{p}'_i), 
	\end{equation}
	in which $d(\bm{p}_i,\bm{p}'_i)$ can be the Manhattan distance between $\bm{p}_i$ to $\bm{p}'_i$ \cite{DBLP:journals/corr/LiY17a}.
	
	DoM reflects how far one solution set $\bm{P}$ needs to move to dominate another solution set $\bm{Q}$. Thus, $DoM(\bm{P}, \bm{Q})$ is always larger than or equal to zero. A small value suggests that $\bm{P}$ and $\bm{Q}$ are close (e.g., one point of $\bm{P}$ that only needs a little movement to dominate $\bm{Q}$), and a large value indicates that $\bm{P}$ performs worst than $\bm{Q}$ (e.g., some points of $\bm{P}$ must be moved further away in an attempt to dominate $\bm{Q}$).
	
	The dominance move indicator is based on the properties of dominance relation among solutions trying to dominate each other, considering all available information (all solutions and objectives). These solutions' efforts scale in a bottom-up manner from the solutions to the set relations. In this sense, the authors claimed that DoM possesses some properties that are enumerated in the following proposition \cite{Li:2019:QES:3320149.3300148}:
	
	\begin{proposition}
		Consider the solution sets $\bm{P}$, $\bm{Q}$, $\bm{R}$ $\subset \Theta$  and the points $\bm{p}$, $\bm{q} \in \Theta$:
		\begin{enumerate}[a)]
			\item $\bm{P}$ =  $\bm{Q}$ $\Longleftrightarrow$ $DoM(\bm{P},\bm{Q}) = DoM(\bm{Q},\bm{P}) = 0$;
			\item $\bm{P} \vartriangleleft \bm{Q}$ ($\bm{P}$ is better than $\bm{Q}$) $\Longleftrightarrow$ $DoM(\bm{P},\bm{Q}) = 0 \land DoM(\bm{Q},\bm{P}) > 0$;
			\item If $\bm{P} \preceq \bm{Q}$, then $DoM(\bm{P},\bm{R})  \leqslant DoM(\bm{Q},\bm{R})$ and  $DoM(\bm{R},\bm{Q})  \leq DoM(\bm{R},\bm{P})$;
			\item $DoM(\bm{P},\bm{Q}) + DoM(\bm{Q},\bm{R}) \geq DoM(\bm{P},\bm{Q} \cup \bm{R})$ and $DoM(\bm{P},\bm{Q}) + DoM(\bm{P},\bm{R}) \geq DoM(\bm{P}, \bm{Q}\cup \bm{R})$.
		\end{enumerate}
		
	\end{proposition}

	The DoM formulation intends to overcome some $\epsilon$-indicators' weaknesses while possessing a similar interpretation. \treviewertwo{The $\epsilon$-indicator measures the maximum of  minimum distance to move one solution set, such that it dominates another solution set.} Then, it must be observed that DoM and $\epsilon$-indicators have the same general concept and lead to the same conclusions among the solution sets. However, DoM presents the following improvements when compared to $\epsilon$-indicators \cite{lopes2020assignment}:
	
	\begin{itemize}
		\item $\epsilon$-indicators are only related to one particular solution and only one objective in the whole solution set;
		\item $\epsilon$-additive is not able to capture differences concerning cardinality of solution sets;
		\item DoM can present greater values than $\epsilon$-indicators. This fact can be explained since DoM takes into account information from all objectives.
	\end{itemize}

	The information loss of $\epsilon$-indicators is critical, particularly considering many-objective scenarios (considering only  the objective on which one solution performs worst relative to another solution and ignoring the difference on the remaining objective values). A very simple example, proposed in \cite{DBLP:journals/corr/LiY17a}, illustrates this information loss: consider two 10-objective solutions, such as $\bm{p}_{1} = \{0, 0, 0, \ldots, 1\}$ and $\bm{q}_{1} = \{1, 1, 1, \ldots, 0\}$. Whereas $\bm{p}_1$ presents a better performance on the first nine objectives and $\bm{q}_1$ only on the last objective, $\epsilon$-additive($\bm{p}_{1}$, $\bm{q}_{1}$) =  $\epsilon$-additive($\bm{q}_{1}$, $\bm{p}_{1}$) = 1. 
	
	Considering DoM's definition again, we aim to obtain \bm{$P'$} from \bm{$P$}  that dominates \bm{$Q$}. In fact, DoM represents the minimum total of moves from  \bm{$P$} $ = \{\bm{p}_1,\bm{p}_2, \ldots, \bm{p}_{|\bm{P}|} \}$ to \bm{$P'$} $ = \{{\bm{p'}_1},{\bm{p'}_2}, \ldots, {\bm{p'}_{|\bm{P}|}} \}$ which is calculated using all available information and not only one objective (as proposed in the additive and multiplicative versions of $\epsilon$-indicator).
	
	\bm{$P'$} represents a set of points that are candidates to dominate \bm{$Q$} with some update in one or more objectives leading to a better distance such as expressed in Equation \eqref{DPQ}. In such way, it should be noted that each $\bm{p'}_i$ with $i \in \{{1, \ldots, |\bm{P}|}\}$ must be generated from {$\bm{p}_i$}, as proposed in \cite{lopes2020assignment}.  
	
	
	\newcommand{\Qs}{\mathcal{Q}}
	
	Let $\mathbb{P}^*(\bm{Q})$ be a set composed of all possible subsets from $\bm{Q}$, excluding the empty set\footnote{It is worthwhile to mention that $\mathbb{P}^*(\bm{Q})$ represents the power set excluding the empty set ($\mathbb{P}^*(\bm{Q}) ~\cup ~\emptyset = \mathbb{P}(\bm{Q})$).}. For determining each element of $\bm{P'}$, $\bm{p'}_i$, it is necessary to associate the corresponding element of $\bm{P}$ to elements of $\mathbb{P}^*(\bm{Q})$. The number of all  possible associations, $\eta$, is the product of $|\bm{P}|$ and $|\mathbb{P}^*(\bm{Q})|$ and is given by
	
	\begin{equation}\label{comb_plinha}
	\begin{array}{lcl}
	\eta & = & |\bm{P}| \underbrace{\sum\limits_{g=1}^{|\bm{Q}|}{\binom{|\bm{Q}|}{g} }}_{|\mathbb{P}^*(\bm{Q})|}, \\ \\
	&=& |\bm{P}| \left({\binom{|\bm{Q}|}{1}} + {\binom{|\bm{Q}|}{2}} + \ldots + {\binom{|\bm{Q}|}{|\bm{Q}|}}\right),  \\ \\
	&=& |\bm{P}|( 2^{|\bm{Q}|} - 1).
	\end{array}
	\end{equation}
	Let $\Qs_s$ be an element of $\mathbb{P}^*(\bm{Q})$ for $s = \{1, \ldots, |\mathbb{P}^*(\bm{Q})|\}$.  Observe that $\bm{p}_{i}$ will be used as a reference point to generate $\bm{p'}_i$. In this way,  $\bm{p'}_i$ may  dominate $\Qs_s$ \cite{lopes2020assignment}. 
	
	
	Regardless of this challenge, the original DoM paper also proposes an exact calculation method for the bi-objective case, which is meticulously  detailed in \cite{Li:2019:QES:3320149.3300148}. For the sake of completeness and clarification, an overview of the algorithm is described next. The algorithm employs the concept of \textit{inward neighbor}:  
	
	\begin{definition}{\textit{Inward neighbor}:}
		Consider $\bm{P}$ a solution set, $\bm{a} \in \bm{P}$, $\bm{b} \in \bm{P}$, and $\bm{b} \ne \bm{a}$.  $ n_{R}(\cdot)$ is a function which determines $\bm{b}$ as the \textit{inward neighbor} of $\bm{a}$, denoted as $\bm{b} = n_{R}({\bm{a}})$, if $\bm{b}$ has the smallest dominance move distance to $\bm{a}$, that is $\bm{b} = \arg\min_{\bm{p} \in \bm{P}/\bm{a}} d(\bm{p},\bm{a})$.  
	\end{definition}
	
	Using the \textit{inward neighbor} concept, the DoM algorithm, as presented in \cite{Li:2019:QES:3320149.3300148},  can be outlined as:
	\begin{list}{}{}
		\item \textit{Step 1:} Remove the dominated points in both \bm{$P$} and \bm{$Q$}, separately. Remove the points of \bm{$Q$} that are dominated by at least one point in \bm{$P$}.
		\item \textit{Step 2:} Denoting  \bm{$R$} $=$ $\bm{P} \cup \bm{Q}$, each point of $\bm{Q}$ in $\bm{R}$ is considered as a subset. For each subset, find its \textit{inward neighbor} $\bm{r} = n_{R}(\bm{q}_j)$ in \bm{$R$}. If the point $\bm{r} \in \bm{P}$, then merge $\bm{r}$ into the subset, otherwise $\bm{r}$ does not belong to the subset, or it is already owned by a subset. At the first case, $\bm{q}_j$ and $\bm{r}$ are merged into the subset. At the second case, there is nothing to do.
		\item \textit{Step 3:} If there exists no point $\bm{q}_j \in \bm{Q}$ such that $\bm{q}_j = n_{R}(n_{R}(\bm{q}_j))$ (i.e., there is a loop between the points) in any subset, then the procedure ends and there is an optimal solution to the case.
		\item \textit{Step 4:} Otherwise, there is a loop in one or more subset, then replace these solutions by their ideal solution (formed of the best of each objective in each solution inside the loop or subset). This leads to a new \bm{$Q$}, then find the \textit{inward neighbor} of such ideal point and subset then.  Return to step 3 until convergence.
	\end{list}
	
	All definitions, theorems, and corollaries to prove the algorithm correctness for the bi-objective case can be seen in \cite{Li:2019:QES:3320149.3300148}. 
	
	Another attempt to solve the problem in the multi-objective and many-objective case was presented in \cite{lopes2020assignment}. The DoM calculation was treated as an assignment problem, and some experiments in three objectives were also presented. However, the proposal has some drawbacks that prevented its use in solutions sets with more than 20 solution candidates. \treviewerone{The assignments necessary to calculate the formulation turns it prohibitive when the number of solutions in the non-dominated set is increased.}. The problem comes from the fact that the \treviewerone{number of dominance moves calculated was combinatorial in nature, detailed in Equation \eqref{comb_plinha}}. Nevertheless, the method presented optimal solutions to some problems, and it is showed that the dimension of the objective space barely influences it. \treviewerone{This study, besides facilitating for the DoM quality indicator to be used in larger solution sets, evaluated the proposed MIP-DoM approach with (i) parametric studies, (ii) time-complexity estimates, (iii) testing on more complex test problems, and (iv) its use as a running quality indicator.}

	\section{The MIP dominance move calculation approach} \label{MIPDOM}
	
	Our goal is to modify the DoM formulation to deal with problems having three or more objective functions. Our DoM calculation proposal is based on the perspective that the problem is, in fact, an instance of an assignment problem with two levels and some restrictions, \treviewerone{but avoiding the combinatorial calculation of the number of dominance moves}, as stated in \cite{lopes2020assignment}. It is considered that to treat the problem, we have to find an assignment of \bm{$P$} to \bm{$Q$} with the restrictions that each {$\bm{q}_{j}$} must be assigned to one {$\bm{p}_{i}$} with the minimum distance. Nevertheless, in classic assignment problems, it is not possible that points in \bm{$P$} change their own features, such as changing positions to alter their distances. In the proposed DoM calculation, this issue is considered.

	A simple example to clarify the situation is given next. Let set \bm{$P$} be  \{(2.0, 2.5), (3.0, 1.9)\} and set \bm{$Q$} be \{(2.2, 2.0), (3.0, 1.5)\}. The \textit{inward neighbors} $\bm{p}_i = n_{R}(\bm{q}_j)$ of points $\bm{q}_{1}$ and $\bm{q}_{2}$  are  $\bm{p}_{1}$ and $\bm{p}_{2}$, respectively. This creates an assignment of \bm{$P$} to \bm{$Q$} with the minimum $DoM(\bm{P},\bm{Q})$. In this sense, a search to find $\bm{p'}_{1}$ and $\bm{p'}_{2}$ should be done, in a way that they must be generated to dominate $\bm{q}_{1}$ and $\bm{q}_{2}$, respectively. Let \bm{$P'$} be \{(2.0, 2.0), (3.0, 1.5)\}, for example, the $DoM(\bm{P},\bm{Q}) = d(\bm{p}_{1}, \bm{p'}_{1}) + d(\bm{p}_{2},\bm{p'}_{2})) = 0.5 + 0.4 = 0.9$. 
	
	\newcommand{\vc}[3]{\overset{#2}{\underset{#3}{#1}}}
	
	The problem can be modeled as a mixed integer programming (MIP) approach and using the DoM definition. \bm{$P$} and \bm{$Q$} are sets of points, with $\bm{p}_i$ points $i \in \{{1,\ldots,|\bm{P}|}\}$ and $\bm{q}_j$ points $j \in \{ {1,\ldots,|\bm{Q}|} \}$, respectively. Each $\bm{p}_i$ is composed of $p_{(i,m)}$ components,  $1 \leq m \leq M$ in which $M$ is the number of objective functions. Analogously, each $\bm{q}_j$ is composed of $q_{(j,m)}$ components,  $1 \leq m \leq M$.
	
	Given $\bm{P}$ and $\bm{Q}$, $\bm{P'}$ is a set of points, in which each $\bm{p'}_i$ is generated from $\bm{p}_i$ after some updates in one or more objectives to dominate a $\Qs_s$. It is important to note that if $\bm{P}$ already dominates $\bm{Q}$, then $\bm{p'}_i = \bm{p}_i$. 
	Observe that \bm{$P'$} must dominate \bm{$Q$}, resulting in a better distance such as expressed in Equation~\eqref{DPQ}.
	
	The proposed MIP model calculates the distance $d(\bm{p}_i,\bm{p'}_i)$. In order to calculate this distance,  a new point, $\bm{\hat{p}}_i$, is obtained 
	and there is strict relation between $\bm{\hat{p}}_i$ and $\bm{p}'_i$. At the end, the MIP-DoM model calculates the final $\bm{p'}_i$ using $\bm{\hat{p}}_i$. 
	
	\begin{figure}[!htb]
		\centering\includegraphics[width=1\linewidth,trim={0cm 4.5cm 0.1cm 0.12cm},clip]{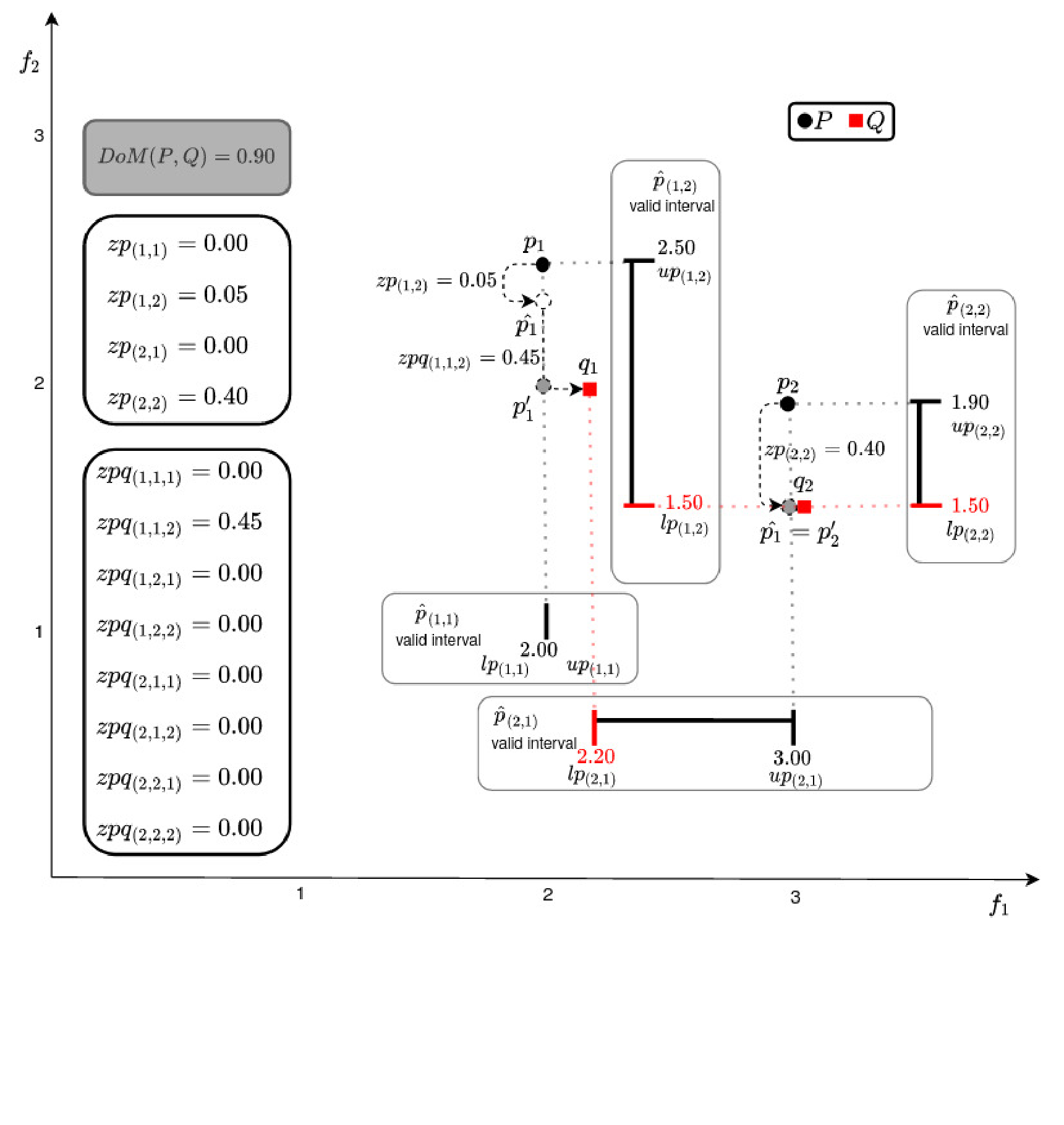}
		\caption{An example that shows up the intuition behinds the MIP-DoM approach. Consider two sets, \bm{$P$} = \{(2.0, 2.5), (3.0, 1.9)\} and \bm{$Q$} = \{(2.2, 2.0), (3.0, 1.5)\}.  The DoM value indicates the total movement for \bm{$P$} to  dominate  \bm{$Q$}. A typical vertical movement,  $zp_{(1,2)}=0.05$ represents an  improvement in $f_{2}$ objective generating $\bm{\hat{p}}_{1}$. However,  $\bm{\hat{p}}_{1}$ does not dominate $\bm{q}_{1}$.  There is still a movement to be done in $f_{2}$ objective,  represented by $zpq_{(1,1,2)}=0.45$, so that $\bm{p'}_{1}$ will dominate $\bm{q}_1$.   Observe that point $\bm{p}_{2}$ does not dominate $\bm{q}_{2}$, so it is vertically moved obtaining $\bm{\hat{p}}_{2}$, in an attempt to dominate $\bm{q}_2$. With this vertical movement, indicated by  $zp_{(2,2)} = 0.40$, $\bm{\hat{p}}_{2}$ dominates $\bm{q}_{2}$, becoming $\bm{p'}_{2}$. The DoM value  is the total distance  using the terms $zp_{(1,2)}, zp_{(2,2)}$, and $zpq_{(1,1,2)}$. Limits on $\bm{p'}$ movements (lower ($lp_{(i,m)}$) and  upper bounds ($up_{(i,m)}$)) are also indicated.}
		\label{fig:sample_model}%
	\end{figure}
	
	Figure \ref{fig:sample_model} illustrates the intuition behind our proposal in a bi-objective problem. The 
	$d(\bm{p}_i, \bm{p'}_i)$ is composed of two terms, $zp_{(i, m)}$ and $zpq_{(i,j,m)}$.  The  summation of all $zp_{(i, m)}$, $\forall \ m \in\{ 1, \ldots, M\}$ results in $d(\bm{p}_i, \bm{\hat{p}}_i)$, Analogously, the summation of all $zpq_{(i, j, m)}$, $\forall \ m \in\{1, \ldots, M\}$ $\forall j \in\{1, \ldots, |\bm{Q}|\}$ results in $d(\bm{\hat{p}_i}, \bm{p'}_i)$.  $\bm{\hat{p}}_i$, indicated as unfilled circles, represents the point after the first movement. For each $\bm{p}_i$ and each \textit{m}-th objective, there are a lower ($lp_{(i, m)}$) and upper ($up_{(i, m)}$) bounds  constraining $\bm{\hat{p}}_i$'s movement. 
	The bounds are generated considering that the $\hat{p}_{(i,m)}$ needs to be smaller or equal to the minimum value in the \bm{$Q$} taking the $m$-th objective into account (lower bound), and cannot be greater than $p_{(i,m)}$ (upper bound). The component $\hat{p}_{(i,m)}$ acts as an extra variable in our MIP model that leverages the calculation. After finding a 
	$\hat{p}_{(i,m)}$ within the corresponding bounds, $zp_{(i, m)}$ represents the extent of movement. 
	In Figure \ref{fig:sample_model}, there are $\bm{\hat{p}}_1$ and $\bm{\hat{p}}_2$, represented as unfilled circles and there are some improvements in $f_2$ objective generating the $zp_{(1,2)}=0.05$ and $zp_{(2,2)}=0.40$. 
	
	\begin{align}
	&\mbox{minimize}~~ \displaystyle \sum\limits_{i=1}^{|\bm{P}|} \displaystyle\sum\limits_{m=1}^{M} zp_{(i,m)}   +  \displaystyle\sum\limits_{i=1}^{|\bm{P}|} \displaystyle\sum\limits_{j=1}^{|\bm{Q}|} \displaystyle\sum\limits_{m=1}^{M} zpq_{(i,j,m)}   \label{model_summation} \\   
	&\mbox{subject to:}   \nonumber \\ 
	&\begin{array}{ll}
	\left\lbrace \begin{aligned}
	& zp_{(i,m)} \geq  0 ,  \\
	& zp_{(i,m)} \geq  p_{(i,m)} xp_{(i)}  - {\hat{p}}_{(i,m)}, 
	\\
	& zp_{(i,m)} \leq  p_{(i,m)} xp_{(i)},  &\\
	\end{aligned}  \right. 
	& \forall i, \forall j, \forall m
	\end{array} \label{restriction_zpio} \\
	& \begin{array}{l}
	\left\lbrace \begin{aligned}  
	& zpq_{(i,j,m)} \geq  0,  \\
	& zpq_{(i,j,m)} \geq  {\hat{p}}_{(i,m)} - q_{(j,m)} 
	-  p_{(i,m)} (1 - xpq_{(i,j)}),  \\
	& M_{(i,j,m)} = Max(0, p_{(i,m)} - q_{(j,m)}), 
	\\
	& zpq_{(i,j,m)} \leq  {\hat{p}}_{(i,m)} - q_{(j,m)} + \cdots & \\ 
	& \quad \cdots + (M_{(i,j,m)} - lp_{(i,m)} - q_{(j,m)}). (1-xpqd_{(i,j,m)}),  & \\
	&zpq_{(i,j,m)} \leq  M_{(i,j,m)} (xpqd_{(i,j,m)}) , &  \\
	\end{aligned} \right.  \\ 
	\hspace{2.2in} \forall i, \forall j, \forall m 
	\end{array} \label{restriction_zpqijo} \\
	& \begin{array}{ll}
	\left\lbrace \begin{aligned} 
	& up_{(i,m)} = p_{(i,m)},   \\
	& lp_{(i,m)} = Min(p_{(i,m)}, Min(q_{(1..|\bm{Q}|, m)})),  
	\\	
	& lp_{(i,m)} \leq  {\hat{p}}_{(i,m)} \leq  up_{(i,m)}, \\
	\end{aligned} \right. 
	\forall i, \forall m 
	\end{array} \label{restriction_lowerandupper} \\
	& \begin{array}{ll}
	\left\lbrace \begin{aligned} 
	& xp_{(i)} \geq  xpq_{(i,j)},     
	\\
	\end{aligned} \right. & \forall i,  \forall j
	\end{array} \label{restriction_xp} \\
	& \begin{array}{ll}
	\left\lbrace \begin{aligned} 
	& xp_{(i)} \leq  \sum\limits_{j=1}^{|\bm{Q}|} xpq_{(i,j)}, & 
	\\
	\end{aligned} \right. &\hspace{.8in}  \forall i 
	\end{array} \label{restriction_sumxp} \\
	& \begin{array}{ll}
	\left\lbrace \begin{aligned} 
	& \sum\limits_{i=1}^{|\bm{P}|}xpq_{(i,j)} = 1,  & 
	\\
	\end{aligned} \right.  & \hspace{1.in} \forall j 
	\end{array} \label{restriction_xp2}\\
	& \begin{array}{ll}
	\left\lbrace \begin{aligned} 
	& xp_{(i)} \in \{0,1\},  &       \\
	& xpq_{(i,j)}\in \{0,1\} , &  
	\\
	& xpqd_{(i,j,m)} \in \{0,1\}, &  \\
	\end{aligned} \right. & \hspace{.8in} \forall i, \forall j, \forall m \end{array} \label{restriction_xpodB} \\
	&\begin{aligned} \hspace{.2in}  \mbox{where, } i = 1, \ldots, |\bm{P}|,j = 1, \ldots, |\bm{Q}|, m = 1, \ldots, M. \nonumber 
	\end{aligned} 
	\end{align}
	
	The $zpq_{(i,j,m)}$ defines the move in $m$-th objective for the $i$-th component of $\hat{p}_{(i,m)}$ to generate the ${p'}_{(i,m)}$. 
	In Figure \ref{fig:sample_model}, the improvement generated by the $\bm{\hat{p}}_{2}$ is good enough to dominate $\bm{q}_{2}$. In that case, $zpq_{(2,2,1)}$ and $zpq_{(2,2,2)}$ are equal to zero, considering that $\bm{\hat{p}}_{2}$ = $\bm{p'}_2$. The same does not happen for $\bm{\hat{p}}_{1}$; there is still a distance such that $\bm{\hat{p}}_{1}$ must be moved to dominate $\bm{q}_{1}$. 
	This distance is indicated by the $zpq_{(1,1,2)} = 0.45$. The final $\bm{p'}_1$ is obtained using $\bm{\hat{p}}_{1}$ and  $zpq_{(1,1,2)}$. 
	
	The distances are calculated using an MIP model expressed from Equation (\ref{model_summation}) to (\ref{restriction_xpodB}).  The model contains continuous \boldz-variables ($zp$ and $zpq$) and binary \boldx-variables (discussed later), and a number of constraint sets, which are combined in an attempt to find the minimum DoM value. 
	The objective function uses \boldz-variables. \treviewerthree{In a practical sense, the inclusion of $zp$ and $zpq$ avoids  equality constraints in the model, transforming an equality constraint into an inequality one}. There is a constraint set related to $zp_{(i,m)}$, and it is expressed in the set of Equations (\ref{restriction_zpio}). Essentially, it must be greater than or equal to zero and it must be less than $p_{(i,m)}$ 
	and, finally, it must be greater than or equal the difference between $p_{(i,m)}$ and ${\hat{p}}_{(i,m)}$.

	

	In the same way, the variable $zpq_{(i,j,m)}$ represents the difference that  each ${\hat{p}}_{(i,m)}$ still have to improve to generate the ${p'}_{(i,m)}$ candidate in an attempt to be less than or equal to some $q_{(j,m)}$. The $zpq_{(i,j,m)}$ can be zero if ${\hat{p}}_{(i,m)}$ is equal to ${p'}_{(i,m)}$, or greater than zero if ${\hat{p}}_{(i,m)}$ is greater than ${p'}_{(i,m)}$. In fact, $zpq_{(i,j,m)}$ variable can be seen as a penalty value for those solutions that still do not dominate a $\Qs_s$.

	The constraint set in Equation (\ref{restriction_zpqijo})  ensures that if ${\hat{p}}_{(i,m)}$ is not less than or equal to $q_{(j,m)}$, then there is a valid value in $zpq_{(i,j,m)}$. It will receive $\max(0,{\hat{p}}_{(i,m)} - q_{(j,m)})$, if the difference to be less than $q_{(j,m)}$ or $0$, otherwise. A binary variable, $xpq_{(i,j)}$ is used to reach out this model condition. Thus, $xpqd_{(i,j,m)}$ assumes 1 to guarantee the maximum value; otherwise, the solution will be infeasible. Moreover, we have used a linearization technique to obtain the maximum function. 

	The ${\hat{p}}_{(i,m)}$ is a continuous variable and the constraint set expressed in Equation (\ref{restriction_lowerandupper}) presents the interval in which ${\hat{p}}_{(i,m)}$ lies within. 
	
	Constraints represented in Equations (\ref{restriction_xp}) and (\ref{restriction_sumxp}) associates the binary variables $xp_{(i)}$ and $xpq_{(i,j)}$ guaranteeing that if $xp_{(i)}$ is `active', at least one $xpq_{(i,j)}$ would be `active' as well. Equation (\ref{restriction_xp}) asserts that a candidate ${\bm{\hat{p}}}_{i}$  coming from $\bm{p}_{i}$ will try to dominate $\Qs_s$, and in this sense, $xp_{(i)}$ will be equal to one and  some  $xpq_{(i,j)}$ will be also equal to one. 
	
	%
	%
	%
	
	Finally, the constraint set in Equation (\ref{restriction_xpodB}) presents all the binary variables used in the model. The $xp_{(i)}$  is used to guarantee that if a ${\hat{p}}_{(i,m)}$ is used in the model, we will know exactly which ${p_{(i,m)}}$ has generated it. In the same way,  $xpq_{(i,j)}$ indicates that  ${\bm{p_{i}}}$ generated ${\bm{\hat{p}}_{i}}$  and is trying to dominate a $\bm{q}_{j}$. These two binary variables are used to guarantee that at least one ${\bm{{p}}}_{i}$ will be associated with a $\bm{q}_{j}$. 
	
	Different MIP solvers can compute the proposed model such as CPLEX \cite{articleKlotz}, GUROBI \cite{gurobi}, and SCIP \cite{GamrathEtal2020OO}, just to name a few. However, there is no guarantee that the problems can be solved optimally in a non-prohibitive computational time. Just to have an idea of the model size regarding the number of variables and constraints, consider the problem having $\bm{P} = \{(2.0, 2.5), (3.0, 1.9)\}$ and $\bm{Q} = \{(2.2, 2.0), (3.1, 1.5)\}$, depicted in Figure \ref{fig:sample_model}. In this case, the model has 16 continuous variables, 22 binary variables, and 60 constraints (including non-negativity constraints). Considering a generic problem, the number of continuous variables, binary variables, and constraints are detailed in the Equations (\ref{expresion_model1}) to (\ref{expresion_model2}), respectively:
	
	%
	%

	\allowdisplaybreaks
	\begin{alignat}{2}
	&\text{\# continuous variables} && =  (2+ {|\bm{Q}|})  ({|\bm{P}|}  M), \label{expresion_model1}\\  
	&\text{\# binary variables} && =  ({|\bm{P}|}  (1 + {|\bm{Q}|} (1 + 2  M)), \\  
	&\text{\# constraints}  &&={|\bm{Q}|} + {|\bm{P}|} (1+ 3 M) + \cdots \nonumber \\  
	& && \cdots + (({|\bm{P}|}  {|\bm{Q}|}) (3+ 4M)). \label{expresion_model2} 
	\end{alignat}

	As an example, considering $M = 5$, $|\bm{P}| = |\bm{Q}| = 200$, there are 202,002 continuous variables, 440,200 binary variables, and 923,400 constraints, making the MIP problem a relatively large-sized optimization problem.

	\section{Experiments}\label{experiments}
	
	\subsection{Bi-objective case}
	
	The first test aims to show how MIP-DoM calculation addresses the four quality indicator facets: convergence, spread, uniformity, and cardinality  \cite{Li:2019:QES:3320149.3300148}. The same  experiments proposed in \cite{DBLP:journals/corr/LiY17a} to solve DoM in the bi-objective case are adopted to verify the model correctness\footnote{We thank the authors for providing the data set.}.
	\begin{figure*}[!htb]
		\centering
		{\includegraphics[scale=0.55]{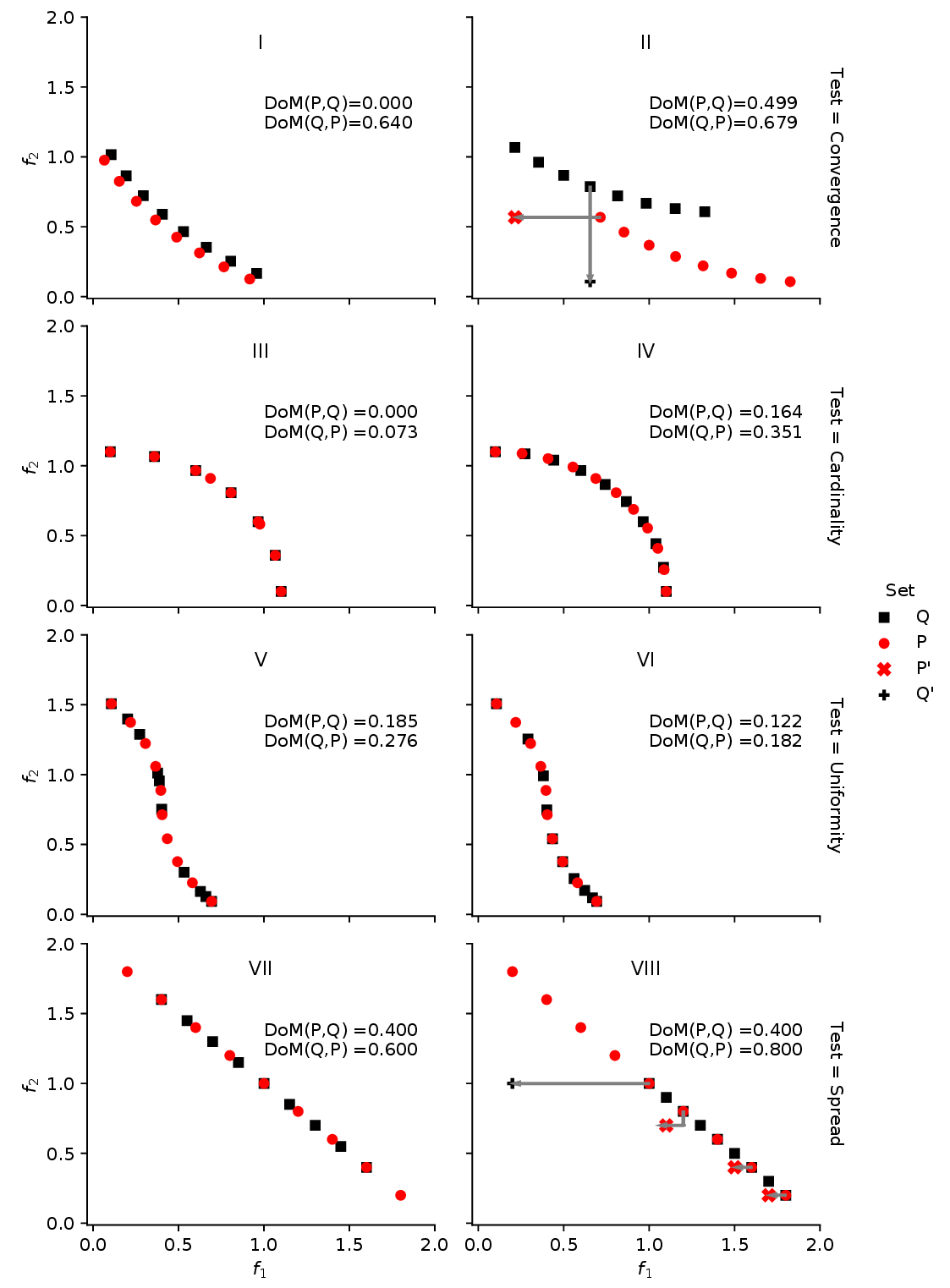} }
		\caption[ArtificialExperiments]{Experiments proposed in \cite{DBLP:journals/corr/LiY17a} to assess the four facets of quality indicators: convergence, cardinality, uniformity, and spread.}
		\label{fig:ArtificialExperiments}%
	\end{figure*}
	
	Figure~\ref{fig:ArtificialExperiments} presents the controlled experiments for assessing the model correctness. Each facet of quality indicators, convergence, cardinality,  uniformity, and spread, is represented in a `row' in Figure \ref{fig:ArtificialExperiments}. Each row has two plots corresponding to the two examples. The examples are slightly different from one another. In each plot, there are $\bm{P}$ and $\bm{Q}$ solution sets.
	
	Convergence is an important factor to reflect Pareto-dominance compliance of sets.  This behavior is shown considering two examples in Figure \ref{fig:ArtificialExperiments},  first row: \textit{test = convergence}. In Examples \textit{I} and \textit{II}, the number of points in $\bm{P}$ and $\bm{Q}$  are equal, but some points of $\bm{P}$ dominate some points of $\bm{Q}$. In plot \textit{I}, $\bm{P}$ completely dominates $\bm{Q}$, hence, dominance move from $\bm{P}$ to $\bm{Q}$, DoM($\bm{P}$,$\bm{Q}$), is zero. In plot \textit{II}, since some points of $\bm{Q}$ are dominated by $\bm{P}$, DoM($\bm{P}$,$\bm{Q}$) is smaller than DoM($\bm{Q}$,$\bm{P}$). 
	There is still a movement sign (gray arrow) in plot \textit{II} indicating the movement for one set to dominate the other. Observe that $\bm{q}_4$ needs to move to $\bm{q'}_4$, thereby dominating $\bm{P}$. On the same way, $\bm{p}_1$ needs to move to $\bm{p'}_1$ to dominate $\bm{Q}$. These are the {\em minimum\/} movement needs to achieve a dominance between the two sets. 
	
	DoM prefers solutions with a different cardinalities in $\bm{P}$ and $\bm{Q}$. In Figure \ref{fig:ArtificialExperiments}, \textit{test = cardinality}, graphics \textit{III} and \textit{IV}, it can be observed that the solution sets have the same convergence, and spread. In graphic \textit{III}, $\bm{Q}$ has one more point than $\bm{P}$. In graphic \textit{IV}, $\bm{P}$ has one more point than $\bm{Q}$. The DoM values of both graphics reflected the cardinality aspect. 
	
	Uniformity indicates the preference for evenly distributed points. The solution sets in Figure \ref{fig:ArtificialExperiments}, \textit{test = uniformity}, presented this feature. The sets have the same convergence, spread, and cardinality. In graphics \textit{V} and \textit{VI}, the set $\bm{P}$ is distributed uniformly, and $\bm{Q}$ has a random distribution. The DoM values show that $\bm{Q}$ needs a bigger move value (DoM) to dominate $\bm{P}$. In graphic \textit{VI}, the density of points in set $\bm{Q}$ increased gradually from bottom to top (considering the $f_1$). Again, a more uniformly distributed set of points needs a smaller DoM to move to a more non-uniformly distributed set of points. 
	
	Finally, DoM must exhibit its preference for solutions having a better spread. A set with better extensity (more extreme points) has a smaller dominance move compared to its competitor. In Figure \ref{fig:ArtificialExperiments}, \textit{test = spread}, considering graphic \textit{VII}, set $\bm{Q}$ was generated by shrinking $\bm{P}$ a little (more concentrated in the middle). In graphic \textit{VIII}, set $\bm{P}$ was distributed uniformly in the range, while $\bm{Q}$ assumed five right bottom points. The graphic \textit{VIII} still presents the $\bm{q}_1$ movement generating $\bm{q'}_1$ to dominate some elements from $\bm{P}$, which were not yet dominated. On the same way $\bm{p}_1$, $\bm{p}_2$ and $\bm{p}_4$, generate $\bm{p'}_1$, $\bm{p'}_2$ and $\bm{p'}_4$ to dominate $\bm{Q}$.
	
	The MIP-DoM approach results are same as the results reported in \cite{DBLP:journals/corr/LiY17a} using the proposed bi-objective algorithm. 
	
	\subsection{Multi- and Many-objective Problems}
	Following the initial and controlled test problems, it is crucial to validate the MIP-DoM model with standard test problem sets. First, a problem set with three objectives is compared with existing indicators, such as the additive $\epsilon$-indicator, hypervolume (HV), IGD+, and also with visual graphics to assess the results. Second, an attempt to solve problems with five, ten and more objectives is made. In all tests, algorithms such as IBEA, MOEA/D, NSGA-III, NSGA-II, and SPEA2 are used to generate the solution sets. It is important to highlight that the goal is to assess the effectiveness of the proposed MIP-DoM approach and not to compare the algorithm's performance, so any other algorithms could have been applied to generate the solution sets. 
	
	
	For each problem set, the population size for the algorithms needs to be initially chosen. We argue that this choice is crucial and is closely related not only to Equation~(\ref{comb_plinha}) but also to the cardinality of the solution set. Considering a good approximation set of the Pareto front, in terms of convergence, spread, and uniformity, the number of non-dominated solutions grows exponentially concerning the dimension of the objective space. 
	In \cite{Sen1998MultipleCD}, the shape of the Pareto front was discussed in the niche size definition. A useful limit to the number of individuals in the population, given by $L = Mr^{M-1}$ is provided, where $r$ is the resolution required or the number of points needed to represent the Pareto front. This expression makes it clear how $|\bm{P}|$, or $|\bm{Q}|$, must be increased as the problem dimension grows, showing an exponential relation between $L$ and $M$. 
	However, other works in many-objective optimization do not strictly follow this rule. In \cite{8027123}, for example, the number of objectives $M$ is 3, 5, and 10, and the population size $L$ is  105, 126, and 275, respectively. In \cite{Yang2019}, an efficient hypervolume calculation is provided and some tests are done with $M$ set to 3, 4, 5 and $L$ between 10 and 200. Likewise, in CEC'2018, a competition on many-objective optimization \cite{cheng2018benchmark} $M$ was chosen to be 5, 10, and 15, and the maximum population size was set to 240.
	
	Since Equation~(\ref{comb_plinha}) plays an  important role in the proposed model (Equations~(\ref{model_summation})  to (\ref{restriction_xpodB})), we have decided to validate DoM using $M = $ 3, 5, 10, and 15 and $|\bm{P}| = |\bm{Q}| = $ 50, 100, 170, and 240 indicating the  final Pareto front approximation size.
	
	All  experiments are done using \textit{Platypus} \cite{Brockhoff:2019:Platypus}, \textit{PyGMO} \cite{PyGMO}, and pymoo \cite{pymoo} to generate the problem sets and to calculate the IGD+ and $\epsilon$-indicator. Hypervolume is calculated using the Walking Fish Group based algorithm \cite{WFGarticle2012} (considering its worst-case complexity as $O(M \times 2^{L})$). 
	The model (Equations (\ref{model_summation})  to (\ref{restriction_xpodB})) is implemented using \textit{Python} and it is solved using \textit{GUROBI} \cite{gurobi} (version 9.0.0 build v9.0.0rc2) running on a Linux 64 bits operational system with 8 CPU's (Intel Xeon E5-2630 v4 2.2GHz) and 32Gb of RAM.
	Some \textit{GUROBI} solver parameters \cite{gurobi} are altered: the minimum $gap = 10^{-6}\%$ , and the $MIPFocus = 3$, which help to improve the best bound during the execution. For all models tested, the optimal solutions are always reached with the established gap.
	
	\subsubsection{Multi-objective Problems}\label{Multi-objective}
	A review of multi-objective benchmark problems is presented in \cite{Huband2011}. To validate our proposed approach, we select some well-known problem test sets in three dimensions, which come from \textit{DTLZ} and \textit{WFG} families. Some test sets are initially chosen based on some characteristics: convexity/concavity, disconnection, multimodality, and degeneracy. DTLZ1, DTLZ2, DTLZ3, and DTLZ7, which are linear, concave/multimodality, concave, and disconnected, respectively, are selected. In the same manner, WFG1, WFG2, WFG3, and WFG9 are also chosen to take into account the properties: convex/mixed, convex/disconnected, linear/degenerate, and concave, respectively. 
	
\begin{figure*}[!htb]
		\centering
		{\includegraphics[scale=0.54]{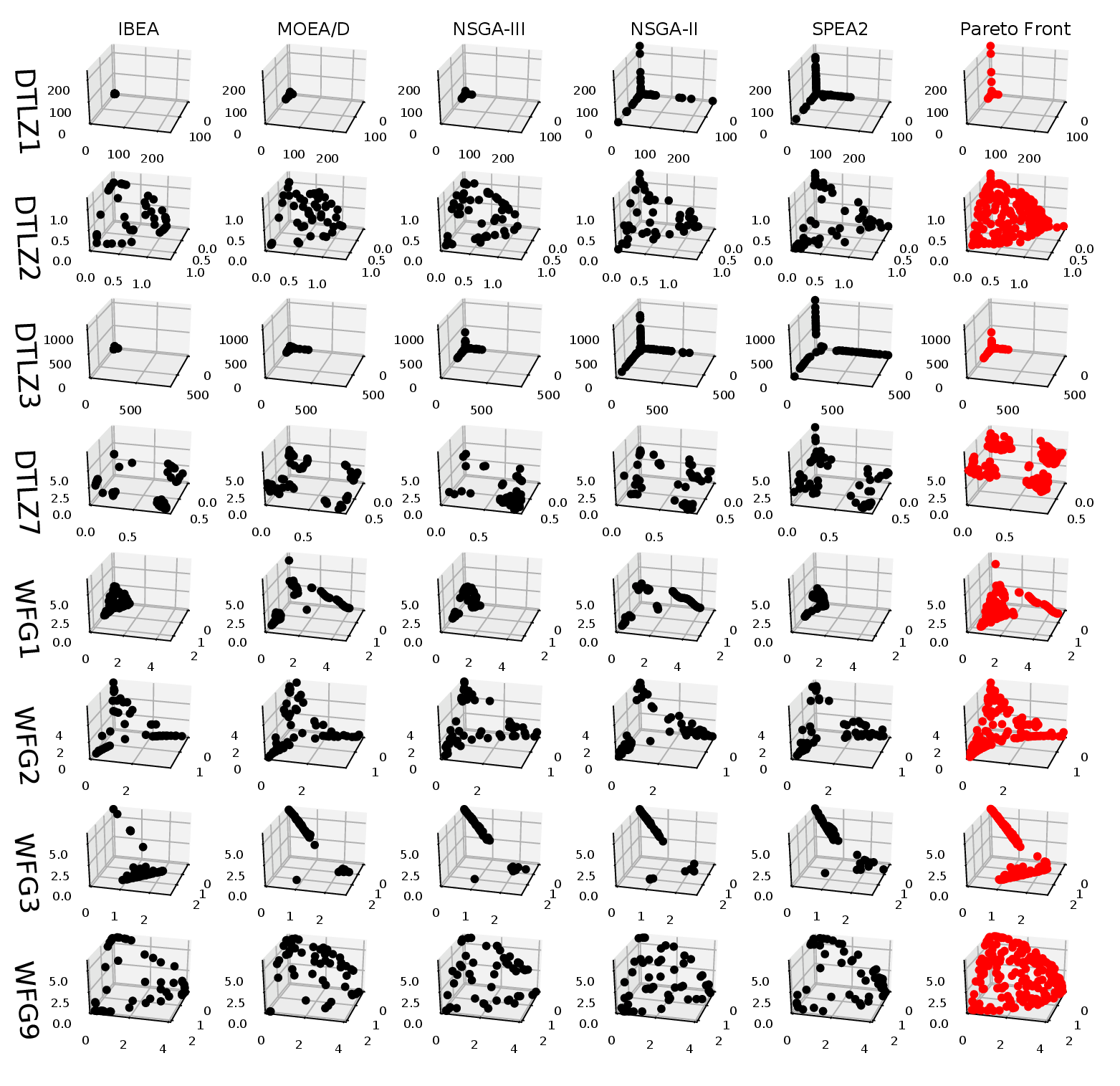} }
		\caption[WFG9]{Solution sets with $|\bm{P}| = |\bm{Q}| = 50$ solutions and three objectives, $M = 3$, generated by IBEA, MOEA/D, NSGA-III, NSGA-II, and SPEA2 algorithms applied to DTLZ1, DTLZ2, DTLZ3, WFG1, WFG2, WFG3 and WFG9 problem sets. The Pareto front is generated by the collecting non-dominated solutions from all algorithms.}
		\label{fig:WFG9_sample}%
	\end{figure*}

	Some algorithms are used to generate the approximated Pareto front, and using the outcomes, the MIP-DoM indicator is compared to other quality indicators. Embedded with a similar purpose of \cite{lopes2020assignment} and \cite{ffb49e1a77a5478fabb03b092c89b2b7}, the NSGA-III and MOEA/D algorithms (Pareto-based and decomposition approach, respectively)  are firstly chosen, and, afterwards, IBEA \cite{Zitzler04indicator-basedselection}, SPEA2 \cite{Zitzler01spea2:improving}, and NSGA-II \cite{Bradford_2018}. All the approximation sets are shown in Figure~\ref{fig:WFG9_sample}.
	
	In our experiments, the maximum number of fitness evaluations is set to 10,000, and each algorithm is executed 21 times. Three unary quality indicators, the additive $\epsilon$-indicator, hypervolume (HV), and the inverted generational distance plus (IGD+), are calculated for each run. It is mandatory to have a reference point or a reference set to calculate the indicators, and this task is a challenging one \cite{Li:2019:QES:3320149.3300148}, \cite{Ishibuchi_Ryo}, but is sometimes provided by the user \cite{Yang2019}. We use the maximum values amongst all algorithm solutions for HV. For additive $\epsilon$-indicator and IGD+, the reference set is a joint Pareto front, which comes from the algorithm's results (it can be viewed in the last column of Figure \ref{fig:WFG9_sample}). 
	

	The MIP-DoM quality measure is a binary indicator. Using MIP-DoM as a unary indicator is still feasible; a straightforward idea is merely using a joint Pareto (created from other algorithm solution sets) or the real Pareto front. The  MIP-DoM is then calculated  using the joint Pareto as a reference set for each problem. We use this approach to facilitate a comparison with some unary indicators, such as HV and IGD+.
	
	Table~\ref{tab:Dom_dtlz1} shows average values of three unary quality indicators, the additive $\epsilon$-indicator, hypervolume (HV), and the inverted generational distance plus (IGD+). MIP-DoM has also been calculated.  In this table, the algorithms are sorted taking into account the ascending order of MIP-DoM results.
	\begin{table}[!htb]
		\centering
		\caption{MIP-DoM($\bm{P}$,$\bm{Q}$) values for the \textit{DTLZ} and \textit{WFG} families for comparison among IBEA, MOEA/D, NSGA-III, NSGA-II, and SPEA2 algorithms. It must be noted that \bm{$P$} is the solution set generated by the algorithm and \textit{\bm{$Q$}} is the non-dominated solution set from all algorithm's results combined. HV, IGD+, and  additive $\epsilon$-indicator are also presented. The algorithms are sorted according to worse performance by MIP-DoM.}
		\label{tab:Dom_dtlz1}
		\begin{tabularx}{\columnwidth}{l l c c c c}
			\hline\noalign{\smallskip}
			\textit{\textbf{Problem}}&\textit{\textbf{Algorithms}}&\textbf{MIP-DoM}&\textbf{HV}&\textbf{IGD+}&\textbf{additive}\\
			&&&&&\textbf{$\epsilon$-indicator}\\
			\noalign{\smallskip}\hline\noalign{\smallskip}
			\textbf{DTLZ1 }
			&\textit{IBEA} &0.312&0.999&0.089&0.001\\
			&\textit{NSGA-III}&0.551&0.999&0.032&0.001\\
			&\textit{MOEA/D}&1.630&0.999&0.052&0.009\\
			&\textit{NSGA-II}&6.422&0.998&0.041&0.027\\
			&\textit{SPEA2}&11.139&0.984&0.089&0.064\\
			\cline{2-6}
			\textbf{DTLZ2}
			&\textit{MOEA/D}&0.970&0.641 &0.065&0.089 \\
			&\textit{IBEA}&1.000&0.632 &0.074&0.130 \\
			&\textit{NSGA-III}&1.004 &0.641&0.065&0.106\\
			&\textit{NSGA-II}&1.013&0.607 &0.076&0.193 \\
			&\textit{SPEA2}&1.021&0.595 &0.082& 0.162 \\
			\cline{2-6}
			\textbf{DTLZ3}
			&\textit{IBEA}&0.0491&0.999 &0.129 &0.000 \\
			&\textit{NSGA-III}&18.652&0.999 &0.017& 0.045 \\
			&\textit{MOEA/D}&25.374&0.998 &0.077 &0.065 \\
			&\textit{NSGA-II}&51.208 &0.984& 0.099&0.142\\
			&\textit{SPEA2}&150.870&0.706& 0.342&0.369 \\
			\cline{2-6}
			\textbf{DTLZ7}
			&\textit{NSGA-III}&1.402&0.198 &0.085&0.175 \\
			&\textit{IBEA} &1.468&0.210 &0.064&0.082\\
			&\textit{MOEA/D}&1.695&0.196 &0.073&0.123\\
			&\textit{NSGA-II}&1.782&0.181 &0.071&0.167\\
			&\textit{SPEA2}&2.184&0.141 &0.112&0.185 \\
			\cline{2-6}
			\textbf{WFG1 }
			&\textit{IBEA} &1.230&0.342&0.192&0.257\\
			&\textit{MOEA/D}&1.557&0.371&0.105&0.243\\
			&\textit{SPEA2}&1.659&0.256&0.289&0.425\\
			&\textit{NSGA-III}&1.822&0.286&0.248&0.326\\
			&\textit{NSGA-II}&1.944&0.343&0.134&0.272\\
			\cline{2-6}
			\textbf{WFG2}
			&\textit{IBEA} &1.503&0.864&0.054&0.110\\
			&\textit{NSGA-III}&1.571&0.824&0.062&0.201\\
			&\textit{MOEA/D}&1.683&0.834&0.057&0.134\\
			&\textit{NSGA-II}&1.687&0.790&0.075&0.184\\
			&\textit{SPEA2}&1.859&0.777&0.080&0.216\\
			\cline{2-6}
			\textbf{WFG3}
			&\textit{IBEA}&1.889&0.410&0.060&0.111\\
			&\textit{NSGA-III}&2.609&0.383&0.069&0.264\\
			&\textit{NSGA-II}&2.630&0.392&0.060&0.241\\
			&\textit{MOEA/D}&2.820&0.375&0.083&0.282\\
			&\textit{SPEA2}&3.178&0.373&0.079&0.211\\
			\cline{2-6}
			\textbf{WFG9}
			&\textit{IBEA}&1.965&0.312&0.085&0.156\\
			&\textit{NSGA-III}&2.194&0.324&0.074&0.095\\
			&\textit{MOEA/D}&2.331&0.290&0.089&0.161\\
			&\textit{NSGA-II}&2.601&0.304&0.074&0.123\\
			&\textit{SPEA2}&2.759&0.289&0.085&0.171\\
			\noalign{\smallskip}\hline
		\end{tabularx}
	\end{table}
	It can be noted from the table that DTLZ1 and DTLZ3 have the HV values inflated by the presence of the dominance resistant solutions \cite{Li:2019:QES:3320149.3300148}, which are non-dominated solutions with a poor value in one objective but with good values in others. 

	For the DTLZ1, the algorithms presenting the best HV, $\epsilon$-indicator, and MIP-DoM values are IBEA and NSGA-III. For the HV indicator, the comparison is difficult due to inflated values. For IGD+, the results indicate NSGA-III and NSGA-II are the best algorithms. For MIP-DoM, the top two algortihms are IBEA and NSGA-III with 0.312 and 0.551, respectively.
	In the DTLZ2 problem, HV, IGD+, $\epsilon$-indicator, and  MIP-DoM indicate MOEA/D as one of the best results. For HV and IGD+, there is a tie between NSGA-III and MOEA/D.
	DTLZ3 results for $\epsilon $-indicator, HV, and MIP-DoM classify IBEA and NSGA-III as the best algorithms. Again, considering HV, due to some extreme points, the values are next to each other. The best algorithms are indicated as  IBEA, NSGA-III, and MOEA/D. For IGD+, there is an indication of NSGA-III and MOEA/D as the best ones.
	Finally, for the DTLZ7 problem set, IBEA is pointed as the best one for HV, IGD+, and $\epsilon$-indicator. However, NSGA-III is the best one, according to MIP-DoM, and the second one considering HV.
	
	The same experiment is done for the \textit{WFG} family. For WFG1 in Table \ref{tab:Dom_dtlz1}, the best algorithm is MOEA/D for HV, IGD+, and $\epsilon$-indicator.  For MIP-DoM, the best algorithms are IBEA and MOEA/D. 
	For WFG2, the best values for  HV, IGD+, and $\epsilon$-indicator and MIP-DoM indicate IBEA as one of the best algorithms. HV, IGD+, and  MIP-DoM ranke these three algorithms as the best ones: IBEA, MOEA/D, and NSGA-III.
	In the same way, for WFG3, IBEA generate the best solution sets for all indicators. Again, there is an agreement amongst them. For IGD+, the best algorithms are IBEA and NSGA-II, presenting a tie between these two algorithms.
	Finally, for the WFG9 problem set, considering HV, IGD+, and additive $\epsilon$-indicator  NSGA-II is the best algorithm. MIP-DoM point out IBEA and NSGA-III as the best ones. For IGD+, there is a tie between NSGA-III and NSGA-II.
	
	\paragraph*{Correlation with Existing Indicators}
	The goal of this paper is to introduce the MIP-DoM as an alternate performance indicator. Next, we calculate Pearson correlation coefficient of the MIP-Dom indicator \treviewerfor{with} HV, IGD+ and additive $\epsilon$-indicator. 
	A Pearson correlation coefficient ($\rho$) is calculated in Table \ref{tab:corr_pear} to assess if there is a linear relation among the indicators. 
	We measure the correlation coefficient using the problem sets individually, the problem set families, and the general correlation among all indicators. To assess the statistical significance of the results, a hypothesis test with a significance level of $0.05$ is carried out. Values which are not statistically significance are indicated as \textit{*} in Table~\ref{tab:corr_pear}.  All values are normalized before the correlation calculation.
	
	\newcolumntype{s}{>{\hsize=.5\hsize}X}
	
	\begin{table}[!htb]
		\caption{The Pearson correlation coefficient ($\rho$) to assess the linear relationship between MIP-DoM and other indicators (HV, IGD+, and $\epsilon$-indicator).}
		\label{tab:corr_pear}
		\centering	
		\begin{threeparttable}
			\begin{tabularx}{\columnwidth}{ l  s s s }
				\hline\noalign{\smallskip}
				&\multicolumn{3}{c}{\textit{\textbf{ MIP DoM }}correlation with}\\\cline{2-4}
				&&&\textbf{\textit{additive}}\\
				\textbf{Problem set}&\textbf{\textit{-HV}}&\textbf{{\textit{IGD+}}}&\textbf{\textit{$\epsilon$-indicator}}\\
				\noalign{\smallskip}\hline\noalign{\smallskip}
				DTLZ1&0.896&0.347*&0.989\\
				DTLZ2&0.770*&0.769*&0.783*\\
				DTLZ3&0.964&0.887&0.998\\
				DTLZ7&0.942 &0.720*&0.502*\\ \cline{2-4}
				\textbf{Combined DTLZ} &\textbf{0.895}&\textbf{0.681}&\textbf{0.818}\\ \hline
				WFG1&0.265*&0.019*&0.256*\\
				WFG2&0.860 &0.815*&0.626*\\
				WFG3&0.945 &0.747*&0.684*\\
				WFG9&0.627*&0.093*&0.226*\\\cline{2-4}
				\textbf{Combined WFG} &\textbf{0.674}&\textbf{0.372*}&\textbf{0.448}\\\cline{2-4}
				\textbf{Combined All} &\textbf{0.784}&\textbf{0.526}&\textbf{0.633}\\
				\noalign{\smallskip}\hline
			\end{tabularx}
			\begin{tablenotes}
				\item[*] We applied a hypothesis test for the correlation coefficient with a significance level of 0.05. The * shows that the calculated p-value is over 0.05, and it is not possible to reject the null hypothesis that $\rho = 0$.
			\end{tablenotes}
		\end{threeparttable}
	\end{table}
	We use the negative HV in the correlation for simplicity since it is the only indicator for which a larger value means better. It is observed that there is a reasonably strong correlation between MIP-DoM and HV, and with additive $\epsilon$-indicator, taking into account all problems. Concerning the DTLZ family, the correlation between MIP-DoM and HV and additive $\epsilon$-indicator is even stronger, $0.90$  and $0.82$, respectively. For the WFG family, the correlations between MIP-DoM and HV and additive $\epsilon$-indicator is different from the DTLZ family. We argue that  WFG1 is responsible for decreasing the correlation values since the correlation between MIP-DoM and HV is high for  WFG2, and WFG3.
	
	In general, the  correlation between MIP-DoM and HV is high for most problems. This gives us confidence of its use as an alternate performance indicator for set-based multi-objective optimization algorithms.
	

	\subsubsection{Many-Objective Problems}
	Next, the goal is to verify if the MIP-DoM approach can be applied in many-objective scenarios. Inspired by the issues raised in the multi-objective experiments and the well-known quality indicators' weaknesses, our motivation to apply DoM in many-objective scenarios can be summed up as: 
	\begin{enumerate}
		\item  HV is hard to exactly calculate in many-objective problems \cite{guerreiro2020hypervolume}. Yet, it is sensitive to extreme points and dominance resistant solutions, such as observed in the last experiment for DTLZ1 and DTLZ3 cases; 
		\item In many-objective problem sets and real cases, it is hard to obtain a Pareto front or reference set that is evenly distributed such as vital to IGD+, for example; The same happens to HV,  in \cite{Ishibuchi_Ryo}, for example, some experiments have shown that a slightly worse point than the nadir point is not always appropriate for problem sets comparison;
		\item Considering the DoM definition and its calculation proposal, it is still relevant to note that there is no information loss in DoM definition. This is an essential feature to a quality indicator, mainly considering problem sets with high number of objective functions and cardinality. 
	\end{enumerate}
	
	In this sense, we would like to observe how the MIP-DoM model behaves in such scenarios, and analyze its characteristics. The same problem sets from the previous experiment are used, and two algorithms are chosen to compare and observe the quality indicator features. 
	We choose MOEA/D and NSGA-III to generate the approximation sets. It is relevant to emphasize that any other algorithm could have been selected, however, the rationale behind this choice is due to  specificity of these algorithms to deal with many-objective problems. The object of our study here is the quality indicator, and the aim is to assess how it behaves with different many-objective test cases.
	
	In the same manner as the CEC'2018 competition \cite{cheng2018benchmark} we establish $L$ = 100, 170, and 240 and $M$ = 3, 5, 10, and 15, respectively. %
	\begin{table}[!htb]
		\centering
		\caption{Using the solution sets generated by MOEA/D and NSGA-III algorithms, MIP-DoM value for the many-objective experiments for \textit{DTLZ}  and \textit{WFG} families.  The number of points in the final non-dominated set is set to $L$ = 100, 170, and 240, and the number of objectives, $M$ = 5, 10, and 15.}
		\label{tab:Dom_many_objective_MT}
		\begin{tabular}{l@{\hspace{2pt}}c@{\hspace{2pt}}c@{\hspace{2pt}}c@{\hspace{2pt}}c@{\hspace{2pt}}c@{\hspace{2pt}}c}
			\hline\noalign{\smallskip}
			&& &&\textbf{\textit{Value}} &\\
			\textit{\textbf{Problem} }&\textit{\textbf{MIP-DoM($\bm{P}$, $\bm{Q}$)} }&&\textit{L=100}&\textit{L=170}&\textit{L=240}   \\
			&&&\textit{M=5}&\textit{M=10}&\textit{M=15}  \\ \cline{1-2} \cline{4-6}
			\textbf{DTLZ1}
			&\textit{(MOEA/D, NSGA-III)} &&1463.461&560.182&594.894\\
			&\textit{(NSGA-III, MOEA/D)}&&3401.332&6315.371&9944.266\\\cline{4-6} 
			\textbf{DTLZ2}
			&\textit{(MOEA/D, NSGA-III)} &&3.943&1.175&1.065\\
			&\textit{(NSGA-III, MOEA/D)}&&3.804&8.972&16.665\\\cline{4-6} 
			\textbf{DTLZ3}
			&\textit{(MOEA/D, NSGA-III)} &&2839.942&1878.871&2040.851\\
			&\textit{(NSGA-III, MOEA/D)}&&6418.077&15532.120&23735.510\\\cline{4-6} 
			\textbf{DTLZ7}
			&\textit{(MOEA/D, NSGA-III)} &&2.502&5.543&4.745\\
			&\textit{(NSGA-III, MOEA/D)}&&1.148&2.948&2.950\\\cline{4-6} 
			\textbf{WFG1}
			&\textit{(MOEA/D, NSGA-III)} &&0.270&0.279&0.124\\
			&\textit{(NSGA-III, MOEA/D)}&&0.464&0.707&0.775\\\cline{4-6} 
			\textbf{WFG2}
			&\textit{(MOEA/D, NSGA-III)} &&2.001&1.179&1.137\\
			&\textit{(NSGA-III, MOEA/D)}&&1.571&2.423&3.665\\\cline{4-6}
			\textbf{WFG3}
			&\textit{(MOEA/D, NSGA-III)} &&4.524&4.678&1.955\\
			&\textit{(NSGA-III, MOEA/D)}&&5.170&14.160&22.981\\\cline{4-6} 
			\textbf{WFG9}
			&\textit{(MOEA/D, NSGA-III)} &&3.183&2.177&0.524\\
			&\textit{(NSGA-III, MOEA/D)}&&4.450&13.583&31.296\\		\noalign{\smallskip}\hline
		\end{tabular}
	\end{table}
	In Table \ref{tab:Dom_many_objective_MT} the MIP-DoM values are presented. The algorithm solution sets are compared in pairs and both directions. For DTLZ1 with $L = 100$ and $M = 5$, the MIP-DoM(MOEA/D, NSGA-III) = 1463.461 and the MIP-DoM(NSGA-III, MOEA/D) = 3401.332, indicating that the approximation set generated by MOEA/D is better than that by NSGA-III. 
	For DTLZ2, while MIP-DoM(NSGA-III, MOEA/D) is smaller than MIP-DoM(MOEA/D, NSGA-III), but for higher objectives, MOEA/D's performance is better. 
	From the table, we observe that MIP-DoM has an agreement about the best algorithm for some problem sets, independent to the parameter $M$ and $L$. In DTLZ1, DTLZ3, WFG1, WFG3, and WFG9, for example, MOEA/D set is better than that in NSGA-III. On the other hand, for DTLZ7, NSGA-III provides lower values. There is a disagreement found in DTLZ2 and WFG2 for $L = 100$ and $M = 5$. Furthermore, for the DTLZ2, for example, the values are similar for both algorithms. 
	However, our goal here is to show that it is possible to use and calculate MIP-DoM regardless the number of points ($L$) in the solution set and the number of objective functions ($M$) and still compare two or more EMaO algorithms. 
	
	
	
	One interesting fact observed in the experiments is related to the number of $\bm{p'}_i$ different from the original $\bm{p}_i$. The minimum number in this case was one, thus the MIP-DoM recommendation is to change only one point. This is always the case behind $\epsilon$-indicators, which exhibit information loss, as previously discussed. On the other side, the maximum number is 9, 8, and 26 in the experiments for $L = 100$ and $M = 5$, $L = 170$ and $M = 10$, and $L = 240$ and $M = 15$, respectively. This fact denotes that MIP-DoM value is more suitable for many-objective scenarios.

	\section{Further Studies with MIP-DoM}\label{sec:further}
	Having compared MIP-DoM indicator with popular existing indicators on multi- and many-objective optimization problems, we now analyze and reveal several other properties of the proposed MIP-DoM indicator. 
	
	\subsection{Computational Time}
	Every MIP-DoM computation requires solving a mixed-integer programming (MIP) problem stated in Equations~(\ref{model_summation}) to (\ref{restriction_xpodB}).
	In cases in which most points of $\bm{Q}$ are dominated by some point in $\bm{P}$ exist, the number of required MIP iterations is small and the computational time to solve the MIP problem is also small. However, in our experiment, the worst-case scenario spends approximately $\sim 208$ hours, for example, using MOEA/D to solve WFG1 with $L = 240$ and $M = 15$. In this case, the pre-processing method is not feasible, and the number of simplex iterations is very large, as well as the number of explored nodes. 
	
\begin{figure}[!htb]
		\centering
		{\includegraphics[scale=0.35]{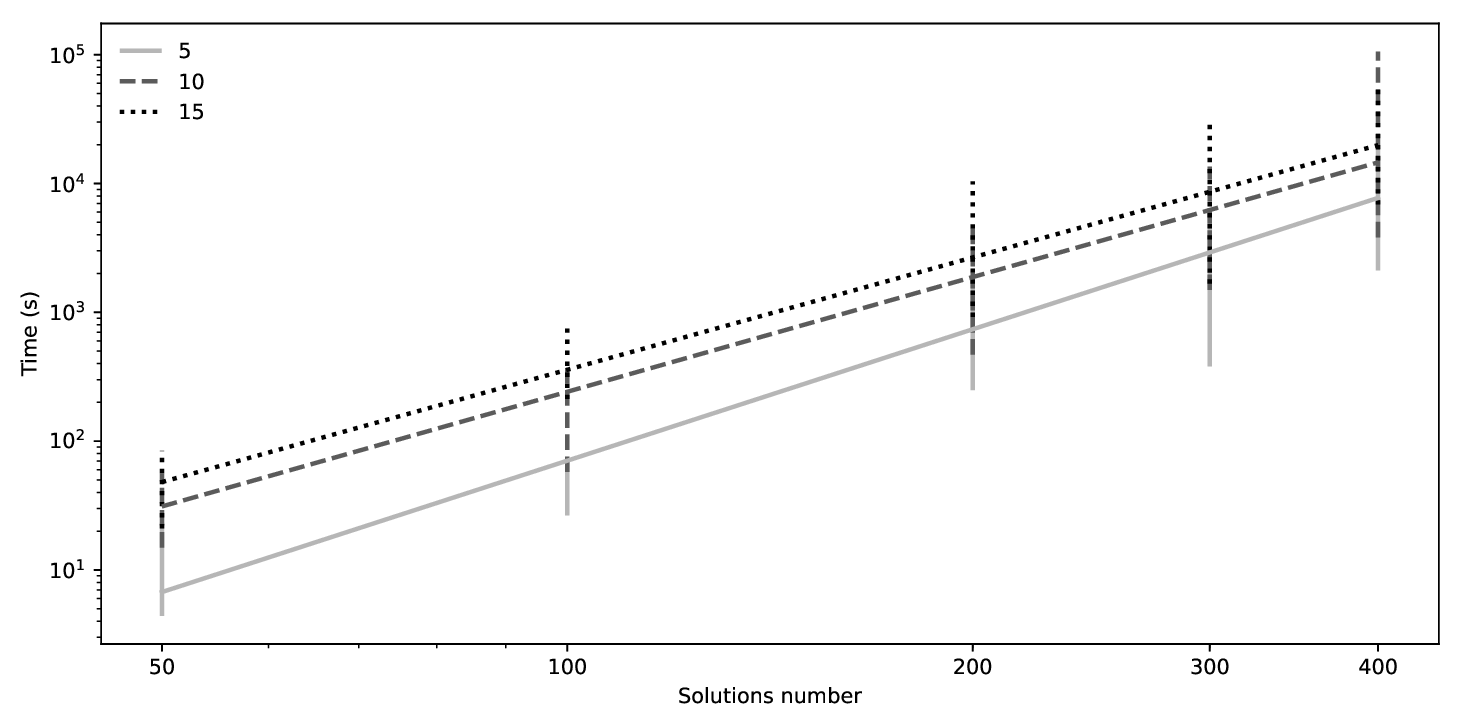} }
		\caption{A polynomial increase in computational time with population size. Fifty experiments are done with $L$ = 50, 100, 200, 300 and 400 solutions using the DTLZ1 problem. The data set is randomly generated, and the objective function is analytically calculated.}
		\label{fig:time_execution_sol}%
	\end{figure}

	The time spent in seconds by the solver is an important aspect to analyze. Figure~\ref{fig:time_execution_sol} presents an experiment using DTLZ1 problem, as an example. We generate 50 random approximation set as $\bm{P}$ and the Pareto front  as $\bm{Q}$ to calculate MIP-DoM($\bm{P}$,$\bm{Q}$) (the distance of movement which a random approximation set needs to dominate the Pareto front). Our intention is to assess how the computational time increases with an increase in number of population size for 5, 10, and 15 objective problems. In Figure ~\ref{fig:time_execution_sol}, the \textit{y} axis is in log scale, and the \textit{x} axis is represented with $L$ = 50, 100, 200, 300 and 400 solutions.
	
	In Figure~\ref{fig:time_execution_sol}, considering M = 5, 10 and 15, we can observe a polynomial time behavior for DTLZ1 ({\color{black}with} $\approx O(L^{3.311})$, $ O(L^{2.888})$ and $O(L^{2.833})$, respectively).

\begin{figure}[!htb]
		\centering
		{\includegraphics[scale=0.35]{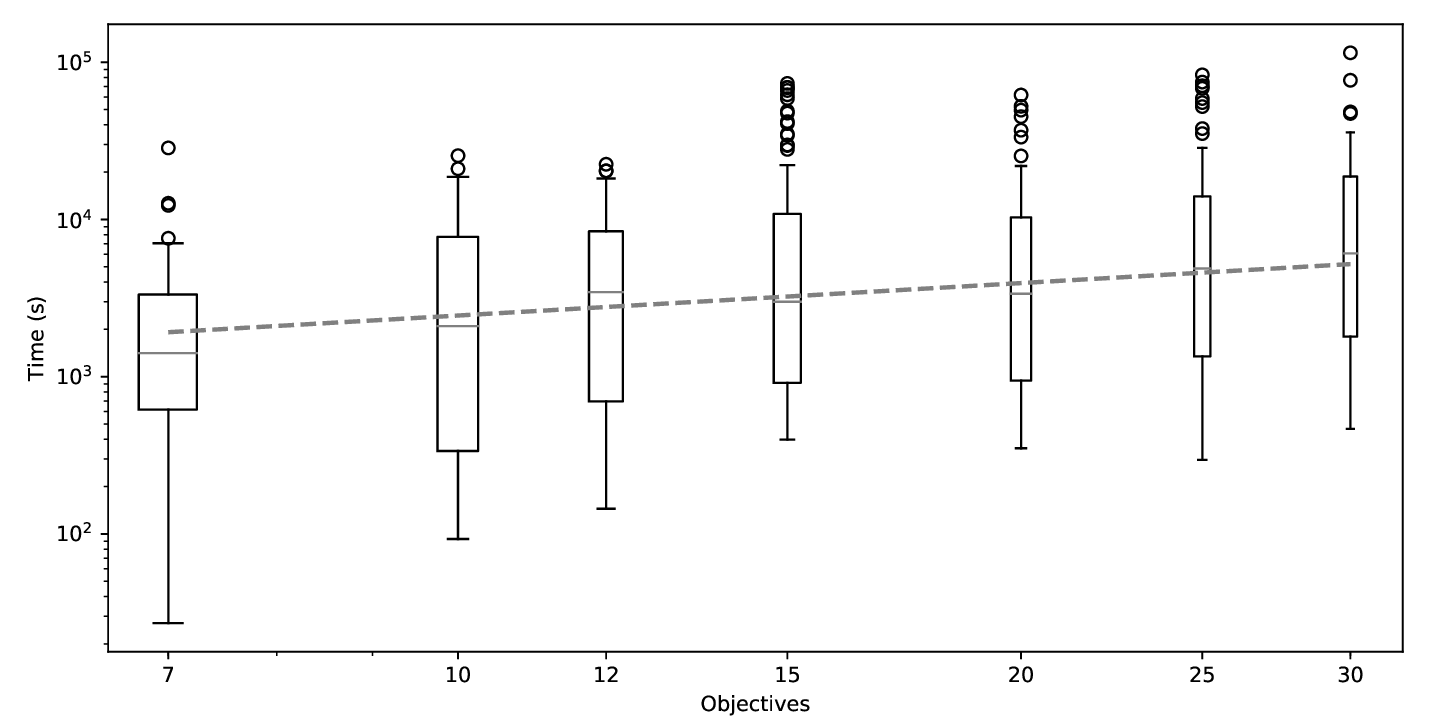} }
		\caption{A boxplot with the time spent to solve the MIP-DoM model as a function of number of objectives. For each box-plot, fifty experiments are done with $L = 200$ using the DTLZ1 problem set definition. The data set is randomly generated, and the objective function is analytically calculated.}
		\label{fig:time_execution}%
	\end{figure}

	Another aspect to verify is how the time behaves if $M$ is increased for many-objective problems. In Figure \ref{fig:time_execution}, using the DTLZ1, values of $M$ = 7, 10, 12, 15, 20, 25 and 30 are adopted and we generate random solutions, and calculate its objective functions using its analytical function. The Pareto front is generated, and our experiments measure the time spent to calculate \textit{MIP-DoM($\bm{P}$, $\bm{Q}$)}, with random approximation set as $\bm{P}$, and the Pareto front as $\bm{Q}$. For each $M$, fifty experiments are run with $L = 200$ solutions in the approximation set. In Figure \ref{fig:time_execution}, both axes are in log scale, representing a linear time behavior ($\approx O(M^{0.686}$). Each box-plot graph is a test set with its associated $M$.
	The figure reveals that the MIP-DoM indicator can be useful for large number of objectives, as it demands a small increase in computational time for an unit increase in number of objectives.
	
	
	A burdensome scenario for MIP-DoM is when there is no a priori dominance between the points in each solution set, and the density of the two solution sets used  are high. In areas with high density, there are many options among $\bm{p}_i$ trying to dominate some $\Qs_s$. The branch-and-bound dynamic becomes more challenging in these situations, combining the $\bm{\hat{p}}$ possibilities, between its lower and upper bound, and the binary variables. This fact must be better investigated and exploited in a future work.

	\subsection{MIP-DoM as a Running Performance Indicator}
	Popular performance metrics (such as, GD, IGD, IGD+, and also HV in some sense) requires the knowledge of true Pareto front so that a reference set can be obtained. This severely restrict their usage in arbitrary problems. But, a recently proposed running performance metric method \cite{julian-deb-runningmetric} can be used for handling an arbitrary problem. At a generation $t$, all non-dominated (ND) solutions from all past generations from the initial population is first collected and the resulting ND set is used as the reference set for metric computation. Then, at generation $t+\Delta t$, the whole process can be repeated ($\Delta t$ can be suitably chosen) and the time varying plot can be made to indicate how the metric changes with generation.  
	
	
	
\begin{figure}[htb]
		\centering
		{\includegraphics[scale=0.35]{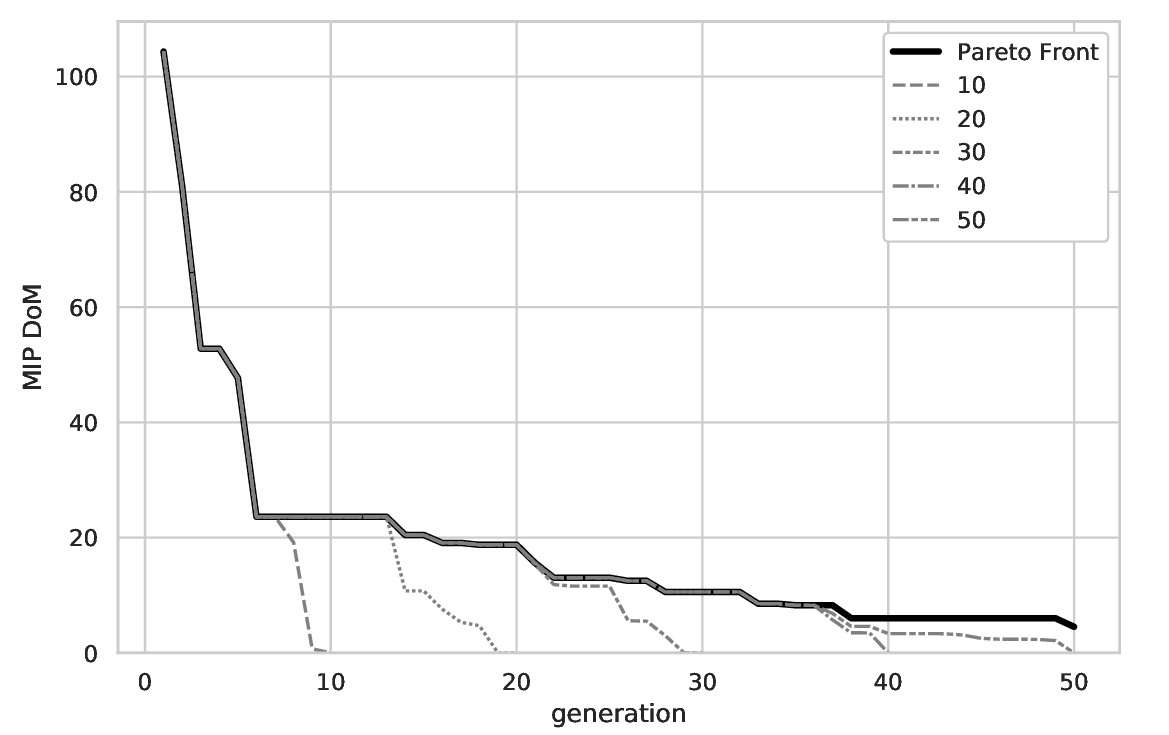} }
		\caption[T]{ MIP-DoM as a running performance quality indicator. The data refers to  DTLZ1 problem set, with two objectives using NSGA-II with 50 generations. We calculate MIP-DoM using  $\bm{P}$ as each generation solution set and  $\bm{Q}$ as solution sets at generations 10, 20, 30, 40, 50 and the final Pareto front.}
		\label{fig:runnning_termination_criteria}%
	\end{figure}

	In \cite{julian-deb-runningmetric}, IGD+ is used as an example of the running quality indicator. The IGD+ indicator demands a reference set, such as the true Pareto front. We show the use of running metric for two purposes: (i) termination criteria in a pair algorithm comparison and (ii) a detailed algorithm behavior view in the quality indicator perspective. 
	Our goal is to verify if MIP-DoM can be used as a running quality indicator, and what are their characteristics. Figure \ref{fig:runnning_termination_criteria} presents six curves. We calculate each one using the solution set from each generation as $\bm{P}$ and, at the generation 10, 20, 30, 40, and 50, solution sets from its generation as $\bm{Q}$ in the MIP-DoM($\bm{P}$, $\bm{Q}$) calculation. The final solid line is computed using the known Pareto front solutions, to demonstrate the behavior of the running indicator. 
	
	Figure 	\ref{fig:runnning_termination_criteria} can be analysed from different manners. First, there is a monotonic decrease in each MIP-DoM curve, indicating that MIP-DoM is able to correctly indicate the monotonic improvement of IGD+ with generations. It is not surprising that each intermediate running indicator curve drops to zero at the pivot generation at which the indicator was calculated. Since the population at this pivot generation is expected to be best compared to earlier generation populations, MIP-DoM($\bm{P_t}$, $\bm{P_t}$) = 0. Second, consider the two consecutive running indicator curve (say at generations 30 and 40). The DoM value calculated at generation 30 using 40-th generation ND set is higher than that calculated using 30-th generation ND set. This indicates that 40-th generation ND set is better than 30-th generation ND set. Third, notice that for up to about 35 generations, DoM calculated using the Pareto front is almost equal to DoM calculated using ND set at generation 50. By checking with a threshold value of MIP-DoM, a termination of an EMO/EMaO run can be adopted.

\begin{figure}[htb]
		\centering
		{\includegraphics[scale=0.35]{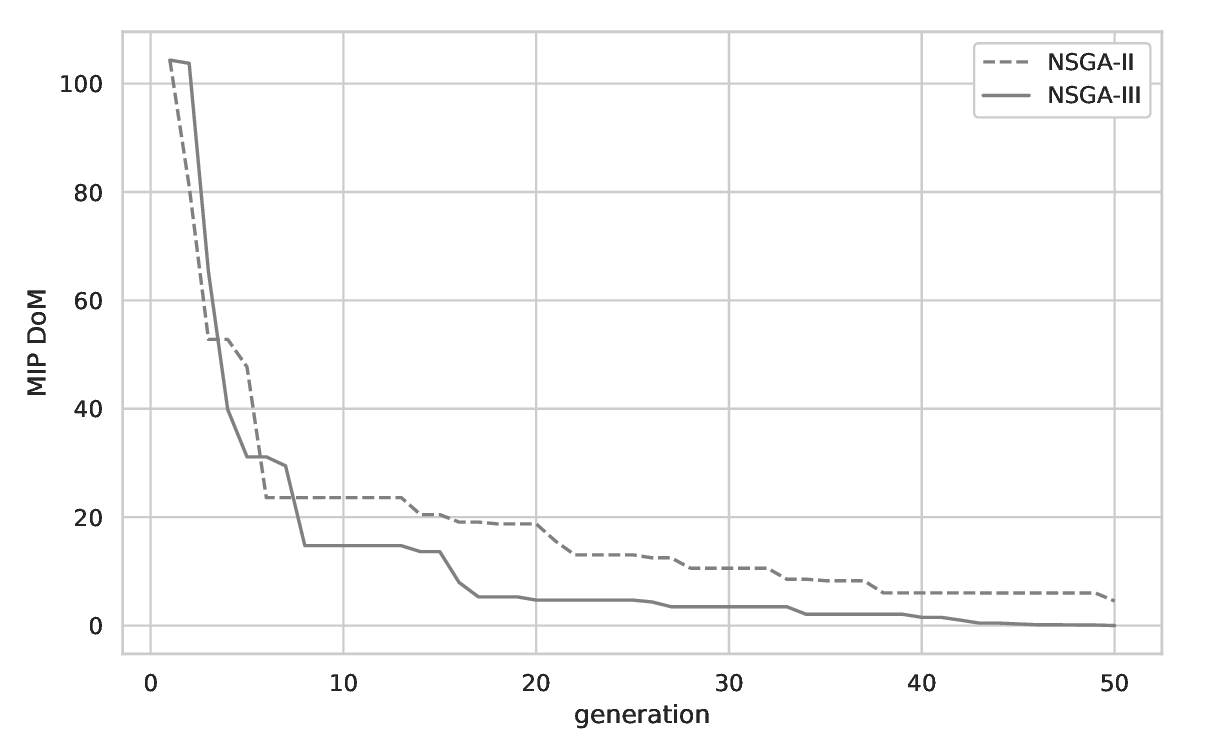} }
		\caption[T]{MIP-DoM running quality indicator applied to DTLZ1 problem set, with two objective functions and 30 solutions in each set. A comparison between NSGA-II and NSGA-III calculating MIP-DoM using $\bm{P}$ as each generation solution set and  $\bm{Q}$ as a joint solution set (non-dominated solutions) from the NSGA-II and NSGA-II 50 generation.}
		\label{fig:runnning_comparison}%
	\end{figure}

	Another use of the running MIP-DoM indicator can come in comparing two or more EMO/EMaO algorithms generation by generation. In Figure \ref{fig:runnning_comparison}, a comparison between NSGA-II and NSGA-III is made. We calculate MIP-DoM(NSGA-II, combined solutions) in a dashed line and MIP-DoM(NSGA-III, combined solutions) in a solid line. The combined solutions represent the non-dominated solutions from NSGA-II and NSGA-III results at generation 50. We use the DTLZ1 with two-objective functions and 30 solutions. Each point in each algorithm curve indicates the MIP-DoM value at the specific generation calculated using the combined solutions. We can observe a similar monotonic drop behavior. In the beginning, until generation 4, NSGA-II presents a better value than NSGA-III. From generation 8 onward, there is a change, and NSGA-III is better than NSGA-II. 
	
	These plots indicate how the MIP-DoM can be employed as a running quality indicator, as termination criteria and as a comparison indicator without the knowledge of the true Pareto solutions.  
	
	\subsection{Parametric Study with Gap in Computing MIP-DoM}
	In the branch-and-bound formulation, some information can be obtained using the function value of the linear programming relaxation and the function value  of the integer program. 
	A valid upper bound on the optimal solution is a proper integer solution with the best value found in the branch-and-bound procedure in a minimization problem. On the opposite side, we also can have a valid lower bound during the branch-and-bound. It comes from the linear programming relaxation, taking the minimum of all current leaf nodes' optimal objective values. The difference between the upper and lower bounds is known as {\em gap}. In general, when the $gap = 0$, we have reached optimality.
	
	The question raised is whether MIP-DoM is affected by a gap variation, such as $gap = 5\%$ or $10\%$? To investigate this issue, we repeat  the same experiment in the multi-objective section, in which the Pearson correlation coefficient between MIP-DoM and some indicators are calculated. The idea is to observe how the gap will influence the MIP-DoM results and how a change affects the correlation coefficient.
	
	We vary the gap parameter, such as posed in our initial question. In Table \ref{tab:times_corr_gap}, not only the original results are presented, but also the results with $gap = 5\%$ and $10\%$ are added. We can observe that the gap parameter introduces a small perturbation in the Pearson correlation coefficient. For example, the HV has a change in the correlation coefficient of less than one percent comparing the values at $gap = 0\%$  and $10\%$. Similar behavior happens to IGD+ and additive $\epsilon$-indicator. 
	
	On the other hand, when we analyze the effect in the model's time, we can observe a clear impact in all cases. In Table \ref{tab:times_corr_gap}, the mean and standard deviation (std deviation) of times for each problem set are presented. The gap has good impacts on the time spent. In some cases, the difference lies in one order of magnitude. 
	
	Additionally, there are differences in the time spent amongst the problem sets. At this experiment, it must be noted that the reference set has different sizes (number of solutions in the joint Pareto), and this influences the time spent by the model, as discussed before. For DTLZ1, DTLZ2, DTLZ3 and DTLZ7 are 79, 207, 93 and 154 solutions, respectively. Considering WFG1, WFG2, WFG3, and WFG9, the values are 104, 124, 151, and 179. Even considering just one problem set, the times spent by MIP-DoM still can show relevant variations among the algorithms. In our experiments IBEA and NSGAIII present the  highest times, considering the DTLZ2 and WFG3.
	
	\begin{table*}[!htb]
		\caption{The first part of the table presents the Pearson correlation coefficient between MIP-DoM and other indicators. The coefficient is re-calculated with the new MIP-DoM value considering $gap = 5\%$ and $10\%$. The second part of the table is focusing on the time spent by the model for the given gap. We show the mean and standard deviation for each problem set and gap.}
		\label{tab:times_corr_gap}
		\begin{adjustbox}{max width=\textwidth}
			\centering	
			\begin{threeparttable}
				\begin{tabularx}{\textwidth}{ l  X X X X X X X X X c X X X X X X }
					\hline\noalign{\smallskip}
					&\multicolumn{9}{c}{\textit{\textbf{ MIP DoM }}correlation with}&&\multicolumn{6}{c}{\textbf{Time spent} (seconds)}\\\cline{2-10} 
					\textbf{Problem set}&\multicolumn{3}{c}{\textbf{\textit{-HV}}}&\multicolumn{3}{c}{\textbf{{\textit{IGD+}}}}&\multicolumn{3}{c}{\textbf{\textit{additive $\epsilon$-indicator}}}&&\multicolumn{3}{c}{mean }&\multicolumn{3}{c}{std deviation}\\
					\cline{2-10}\cline{12-17}
					&\multicolumn{3}{c}{\textbf{\textit{$gap$}}}&\multicolumn{3}{c}{\textbf{\textit{$gap$}}}&\multicolumn{3}{c}{\textbf{\textit{$gap$}}}&&\multicolumn{3}{c}{\textbf{{\textit{$gap$}}}}&\multicolumn{3}{c}{\textbf{{\textit{$gap$}}}}\\
					&${0\%}^{1}$ &$5\%$ &$10\%$ &${0\%}^{1}$ &$5\%$ &$10\%$ &${0\%}^{1}$ &$5\%$ &$10\%$ &&${0\%}^{1}$&$5\%$&$10\%$&${0\%}^{1}$&$5\%$&$10$\%\\
					\noalign{\smallskip}\hline\noalign{\smallskip}
					DTLZ1&0.896&0.896&0.896&0.347*&0.347*&0.347*&0.989&0.989&0.989&&2.4e+01&2.5e+01&1.4e+01&1.2e+01&2.9e+01&1.2e+01\\
					DTLZ2&0.770*&0.773*&0.770*&0.769*&0.763*&0.769*&0.783*&0.784*&0.783*&&2.8e+05&2.0e+05&1.3e+05&3.9e+05&3.3e+05&2.6e+05\\
					DTLZ3&0.964&0.964&0.963&0.887&0.887&0.886&0.998&0.998&0.998&&3.8e+01&2.8e+01&1.6e+01&4.7e+01&2.9e+01&1.8e+01\\
					DTLZ7&0.942&0.942&0.942 &0.720* &0.720* &0.720*&0.502*&0.502*&0.502*&&1.6e+04&3.1e+03&9.9e+02&2.8e+04&3.3e+03&8.3e+02\\ \cline{2-10}\cline{12-17}
					\textbf{Combined DTLZ} &\textbf{0.895}&\textbf{0.894}&\textbf{0.895}&\textbf{0.681}&\textbf{0.679}&\textbf{0.681}&\textbf{0.818}&\textbf{0.818}&\textbf{0.818}&&\textbf{7.5e+04}&\textbf{5.1e+04}&\textbf{3.3e+04}&\textbf{2.2e+05}&\textbf{1.7e+05}&\textbf{1.3e+05}\\ \cline{1-10}\cline{12-17}
					WFG1&0.265*&0.250*&0.242*&0.019*&0.003*&0.002*&0.256*&0.242*&0.224*&&3.3e+02&1.6e+02&8.7e+01&3.6e+02&1.2e+02&6.1e+01\\
					WFG2&0.860&0.877*&0.877*&0.815*&0.833*&0.833*&0.626*&0.631*&0.631*&&3.6e+02&2.0e+02&1.2e+02&3.0e+02&1.3e+02&7.4e+01\\
					WFG3&0.945 &0.945&0.946&0.747*&0.747*&0.747*&0.684*&0.684*&0.684*&&1.6e+04&2.2e+03&7.4e+02&1.7e+04&1.3e+03&2.8e+02\\
					WFG9&0.627*&0.627*&0.649*&0.093*&0.093*&0.009*&0.226*&0.226*&0.333*&&9.9e+02&4.7e+02&2.4e+02&1.1e+03&3.4e+02&1.2e+02\\\cline{2-10}\cline{12-17}
					\textbf{Combined WFG} &\textbf{0.674}&\textbf{0.675}&\textbf{0.678}&\textbf{0.372*}&\textbf{0.372*}&\textbf{0.398*}&\textbf{0.448}&\textbf{0.446}&\textbf{0.468}&&\textbf{4.4e+03}&\textbf{7.5e+02}&\textbf{3.0e+02}&\textbf{1.1e+04}&\textbf{1.1e+03}&\textbf{3.1e+02}\\\cline{1-10}\cline{12-17}
					\textbf{Combined All} &\textbf{0.784}&\textbf{0.784}&\textbf{0.786}&\textbf{0.526}&\textbf{0.526}&\textbf{0.539}&\textbf{0.633}&\textbf{0.632}&\textbf{0.643}&&\textbf{4.0e+04}&\textbf{2.6e+04}&\textbf{1.7e+04}&\textbf{1.6e+05}&\textbf{1.2e+05}&\textbf{9.2e+04}\\\cline{1-17}
				\end{tabularx}
				\begin{tablenotes}
					\item[$^1$]In fact, the ${0\%}$ means a $gap = 10^{-6}\%$.
					\item[*] We use a hypothesis test for the correlation coefficient with a significance level of 0.05. The * shows that the calculated p-value is over 0.05, and it is not possible to reject the null hypothesis that $\rho = 0$.
				\end{tablenotes}
			\end{threeparttable}
		\end{adjustbox}
	\end{table*}
	
	The gap information can help in the binary quality indicator context. If one is interested in a provably optimal solution, it is necessary to wait until the gap reaches zero. Usually, in the quality indicator context, optimality is not an issue and  ``$\bm{P}$ is better than $\bm{Q}$'' is the only question to be answered. Moreover, the faster the answer, the better. Motivated by these issues, different gaps are analysed and the time spent to decide whether  ``$\bm{P}$ is better than $\bm{Q}$'' is computed in each case. Considering Table  \ref{tab:Dom_many_objective_MT} in the many-objective experiments, some high times are also found, for example, DTLZ7 and WFG1 for MOEA/D with $L = 240$ and $M = 15$. For these cases, and in the context of a binary quality indicator, it is not mandatory to obtain an optimal solution, providing the statement is valid. Taking, for example, DTLZ7 for NSGA-III and MOEA/D with $L = 240$ and $M = 15$, it took $\sim 1,818$ seconds to calculate MIP-DoM(NSGA-III, MOEA/D) = 4.475. Similarly, the solver took $\sim 198,795$ seconds to calculate  MIP-DoM(MOEA/D, NSGA-III) = 2.950. However, the solver presented a MIP-DoM valid interval search between 6.188 and 3.101, with $gap = 50\%$ at 98,294 seconds. With this interval, it is possible to assert that NSGA-III is better than MOEA/D, even before the branch-and-bound conclusion.

	\section{Conclusions}\label{conclusion}
	
	The DoM binary quality indicator considers the minimum move of one set to dominate the other set, \treviewertwo{meaning that every element of the second set is either dominated or identical to at least one element of the first set}. 
	The indicator is Pareto compliant and 
	does not demand any parameters or reference sets. The handicap about DoM is its requirement to solve a mixed-integer programming (MIP) problem, making it apparently time consuming to be applied to problems with many objectives and a large population size. 
	
	In this paper, we have provided a detailed MIP formulation of the DoM calculation indicating how the number of variables increase with the number of objectives and population size. The results of our approach have been verified with the original proposal on artificial bi-objective examples. The results have also make the sensitivity of the DoM indicator on dominance, diversity, cardinality, and other issues related to two non-dominated set comparison. 
	
	We have also reported that the MIP-DoM indicator values are highly correlated with other common performance indicators, such as hypervolume (HV), IGD+ and additive $\epsilon$-indicator. The proposed indicator is more correlated with the HV indicator. MIP-DoM indicator has been successfully applied to indicate the performance on three to {\color{black}30}-objective test problems from DTLZ and WFG test suites with population sizes varying from 30 to 400, making an extensive evaluation of the MIP-DoM indicator. 
	
	
	Based on this extensive study, we have observed that MIP-DoM brings following advantages as a binary quality indicator: 
	\begin{itemize}
		\item It does not require any pre-defined set of points, such as, a reference point or a reference set;
		\item It does not demand any normalization of the objective functions;
		\item It is not affected by dominance resistant solutions, unlike that in HV;
		\item It considers all the solutions of both sets being used. There is no lack of information. 
		\item Presumably, MIP-DoM is compatible in indicating four performance facets between two sets: convergence, spread, uniformity, and  cardinality;
		\item It appears to have a highly Pearson correlation with HV;
		\item It presents a monotonic decrease in its value, when the first set ($\bm{Q}$) is fixed and the second set ($\bm{P}$) approaches towards the Pareto front; 
		
	\end{itemize}
	
	However, MIP-DoM  quality indicator still present some drawbacks:
	\begin{itemize}
		\item MIP-DoM time calculation is polynomial to number of variables and objectives, and is more affected by the number of solutions in the set;
		\item Although it can be used as a running quality indicator and presents a correct ranking among curves, in some cases, the difference in MIP-DoM values between two consecutive generations are in decimal places. \treviewerthree{Since MIP-DoM indicates the minimum Minkowski distance move to make one set to dominate another, for two close sets the MIP-DoM can be quite small, compared to the change in hypervolume or IGD metrics};
		\item It demands an efficient MIP solver in its calculation.
	\end{itemize}
	
	Despite this extensive study on analyzing properties of the MIP-DoM indicator, there are a number of future studies which are worth pursuing. For some experiments in our study, we observed a high variability in the computational time from generation to generation for the same problem size and population size. It was observed that this fact was inherent to the distribution or density, and some inner characteristics involving the two solution sets, $\bm{P}$ and $\bm{Q}$. This issue is something that deserves to be better investigated in order to improve the MIP model, creating, for example, new constraints in the MIP model.
	
	In the MIP approach for DoM, other directions also deserve to be investigated: i) How to efficiently calculate DoM using MIP and exploiting some inherent solution set features, such as solution set density; ii) How to define a priori the minimum number of $\bm{p}_i\ne \bm{p'}_i$ and what benefits it can bring to the MIP model; iii) How to improve the convergence time controlling the ill-conditioning of the coefficient matrix induced by the problem formulation presented in \ref{MIPDOM}. 
	Such questions have been posed to emphasize possible future research directions. 
	
	
	Finally, besides its use as quality indicator, MIP-DoM generates a new solution set which dominates another set, $\bm{P'}$. This information can be used within genetic operators to create better solutions for a faster convergence. All these are interesting extensions, nevertheless, we have revealed useful features of the MIP-DoM indicator which should encourage EMO/EMaO researchers to pay more attention to it. 
	
	\section*{Acknowledgements}
	Flávio Martins and Elizabeth Wanner would like to thank the support from the Brazilian funding agencies: CAPES, FAPEMIG and CNPq. Flávio Martins also acknowledges the support from Michigan State University (MSU) (under Koenig Endowed Chair grant of Prof. Kalyanmoy Deb) for his visit to MSU.
	

	
	
	\bibliographystyle{IEEEtran}
	\bibliography{IEEEabrv,mip_dom_ieee_v02}

	%
	
	\begin{IEEEbiography}[{\includegraphics[width=1in,height=1.25in,clip,keepaspectratio]{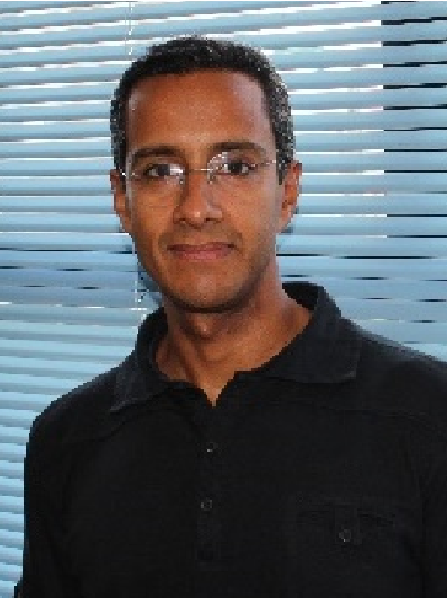}}]{Cláudio Lúcio V. Lopes}
		is a computer scientist and Ph.D. student at Computational and Mathematical Modelling program at CEFET-MG, Belo Horizonte, Brazil. He is a teaching assistant at PUC-MINAS, CTO, and founder at A3Data.  His research interests include evolutionary computation, multi and many-objective optimization, quality indicators, machine learning and artificial intelligence.
	\end{IEEEbiography}
	
	\begin{IEEEbiography}[{\includegraphics[width=1in,height=1.25in,clip,keepaspectratio]{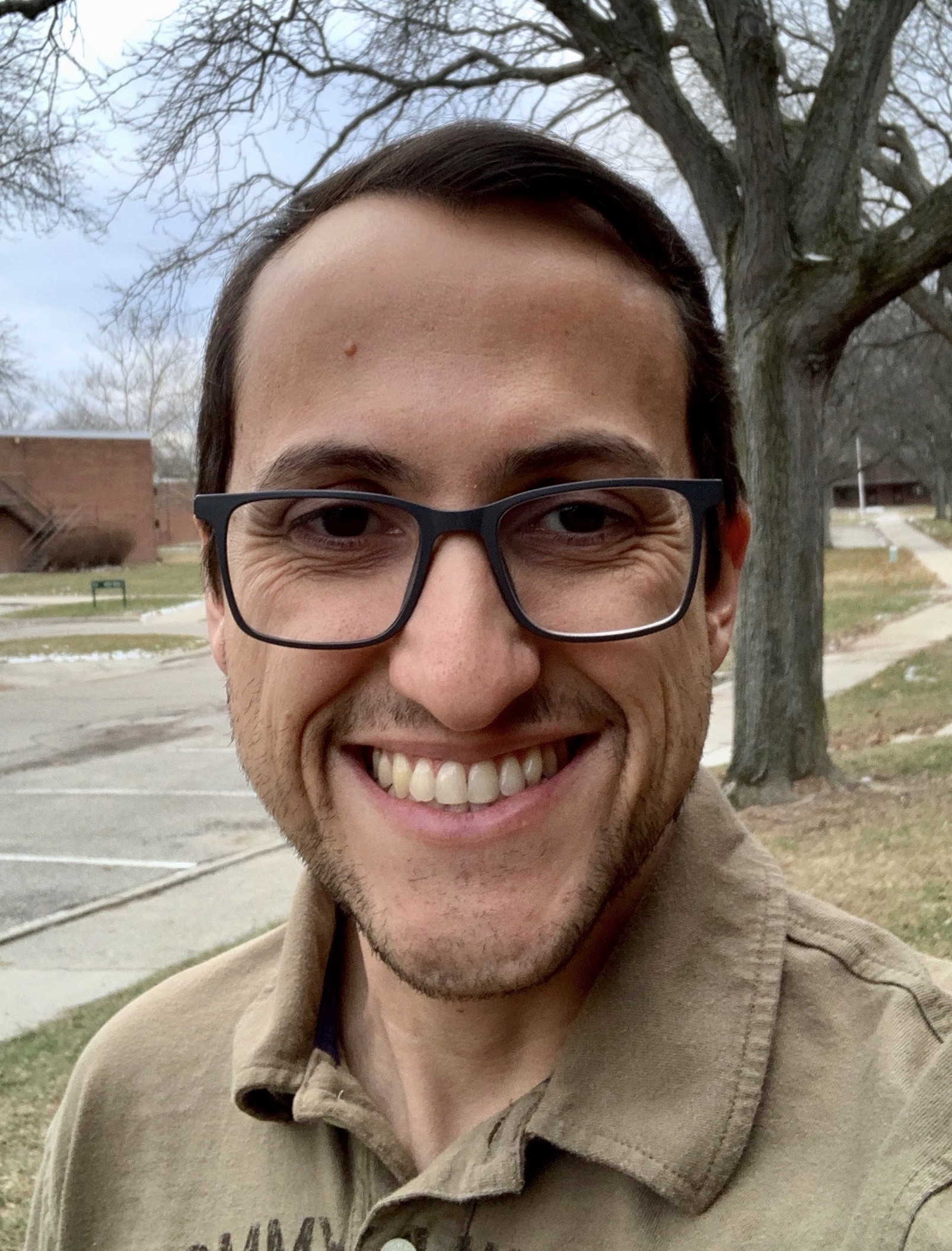}}]{Flávio V. Cruzeiro Martins} received the B.S. degree in Computer Science (2007), M.Sc. degree (2009), and a Ph.D. degree (2012) in Electrical Engineering from Universidade Federal de Minas Gerais. He is currently an assistant professor with the Computer Engineering Department at the Centro Federal de Educação Tecnológica de Minas Gerais, Belo Horizonte, Brazil. His current research interests include evolutionary algorithms, multi-objective and combinatorial optimization, and optimization applied in real-world problems.
	\end{IEEEbiography}
	
	\begin{IEEEbiography}[{\includegraphics[width=1in,height=1.25in,clip,keepaspectratio]{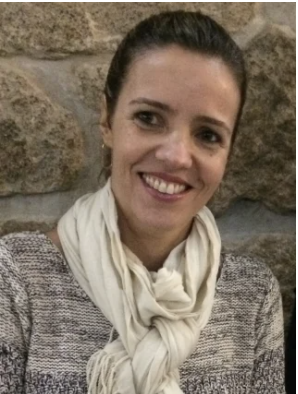}}] {Elizabeth F. Wanner} received the B.S. degree in mathematics (1994) and the M.Sc. degree in mathematics (2002), and a Ph.D. degree in electrical engineering (2006), from the Universidade Federal de Minas Gerais. She is an Associate Professor with the Department of Computer Engineering, Centro Federal de Educação Tecnológica de Minas Gerais, Belo Horizonte, Brazil. Her current research interests include evolutionary computation, global optimization, constraint handling techniques for evolutionary algorithms, and multiobjective optimization.
	\end{IEEEbiography}
	
	\vspace{-1cm}
	\begin{IEEEbiography}[{\includegraphics[width=1in,height=1.25in,clip,keepaspectratio]{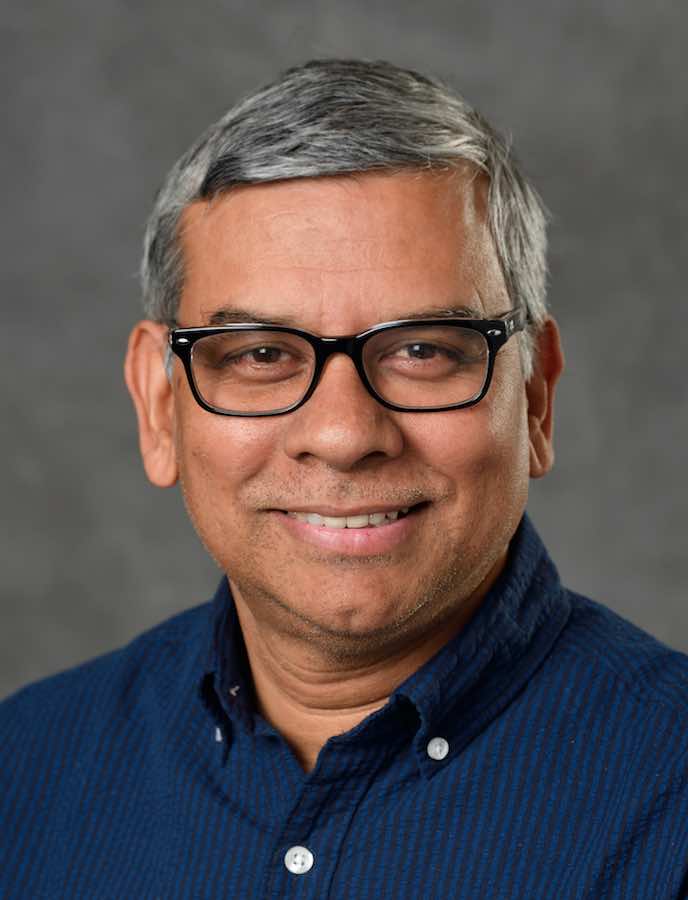}}]{Kalyanmoy Deb}
		is Fellow, IEEE and the Koenig Endowed Chair Professor with the Department of Electrical and Computer Engineering, Michigan State University, East Lansing, Michigan, USA. He received his Bachelor’s degree in Mechanical Engineering from IIT Kharagpur in India, and his Master’s and Ph.D. degrees from the University of Alabama, Tuscaloosa, USA, in 1989 and 1991, respectively. He is largely known for his seminal research in evolutionary multi-criterion optimization. He has published 549 international journal and conference research papers to date. His current research interests include evolutionary optimization and its application in design, modeling, AI, and machine learning. He is one of the top cited EC researchers with more than 149,000 Google Scholar citations and a h-index of 122.
	\end{IEEEbiography}

	
	

\end{document}